\documentclass[final,5p,times]{elsarticle}

\journal{Robotics and Autonomous Systems}

\usepackage{amssymb}
\usepackage{bm}
\usepackage{booktabs}
\usepackage{glossaries}
\usepackage[bookmarks=true]{hyperref}
\usepackage{multirow}
\usepackage[binary-units=true]{siunitx}
\usepackage{subfigure}

\newacronym{C-space}{$\mathcal{C}$-space}{configuration space}
\newacronym{CoM}{CoM}{Center of Mass}
\newacronym{DDP}{DDP}{Differential Dynamic Programming}
\newacronym{DoF}{DoF}{Degrees of Freedom}
\newacronym{EoM}{EoM}{Equation of Motion}
\newacronym{FK}{FK}{Forward Kinematics}
\newacronym{GIW}{GIW}{Gravito-Inertial Wrench}
\newacronym{IK}{IK}{Inverse Kinematics}
\newacronym{IP}{IP}{Interior-Point method}
\newacronym{LP}{LP}{Linear Programming}
\newacronym{NLP}{NLP}{Nonlinear Programming}
\newacronym{QP}{QP}{Quadratic Programming}
\newacronym{SDP}{SDP}{Semidefinite Programming}
\newacronym{SQP}{SQP}{Sequential Quadratic Programming}
\newacronym{TTD}{TTD}{truss topology design}

\graphicspath{{figures/}}
\hypersetup{hidelinks}

\hypersetup{
    pdfauthor   = {Henrique Ferrolho},%
    pdftitle    = {Residual Force Polytope: Admissible Task-Space Forces of Dynamic Trajectories},%
    pdfsubject  = {Robotics},%
    pdfkeywords = {Robots, Trajectory Optimization, Robustness, Polytopes}%
}

\DeclareMathOperator*{\argmin}{\operatornamewithlimits{\arg\!\min}}

\begin{document}

\begin{frontmatter}

    \title{Residual Force Polytope: Admissible Task-Space Forces of Dynamic Trajectories\tnoteref{title_note}}
    \tnotetext[title_note]{This research is supported by The Alan Turing Institute, the EPSRC UK RAI Hub for Offshore Robotics for Certification of Assets (ORCA, EP/R026173/1), EU H2020 project Memory of Motion (MEMMO, 780684), EU H2020 project Enhancing Healthcare with Assistive Robotic Mobile Manipulation (HARMONY, 9911237), and the EPSRC as part of the Centre for Doctoral Training in Robotics and Autonomous Systems at Heriot-Watt University and The University of Edinburgh (EP/L016834/1).}

    \author[address_edinburgh]{Henrique Ferrolho\corref{corresponding_author}}
    \cortext[corresponding_author]{Corresponding author}
    \ead{henrique.ferrolho@ed.ac.uk}

    \author[address_oxford]{Wolfgang Merkt}
    \author[address_edinburgh]{Carlo Tiseo}
    \author[address_edinburgh]{Sethu Vijayakumar}

    \address[address_edinburgh]{Edinburgh Centre for Robotics, School of Informatics, University of Edinburgh, Edinburgh EH8 9AB, United Kingdom}
    \address[address_oxford]{Oxford Robotics Institute, University of Oxford, Oxford OX2 6NN, United Kingdom}

    \begin{abstract}
        We propose a representation for the set of forces a robot can counteract using full system dynamics: the \textit{residual force polytope}.
        Given the nominal torques required by a dynamic motion, this representation models the forces which can be sustained without interfering with that motion.
        The residual force polytope can be used to analyze and compare the set of admissible forces of different trajectories, but it can also be used to define metrics for solving optimization problems, such as in trajectory optimization or system design.
        We demonstrate how such a metric can be applied to trajectory optimization and compare it against other objective functions typically used.
        Our results show that the trajectories computed by optimizing objectives defined as functions of the residual force polytope are more robust to unknown external disturbances.
        The computational cost of these metrics is relatively high and not compatible with the short planning times required by online methods, but they are acceptable for planning motions offline.
    \end{abstract}


\end{frontmatter}

\section{Introduction}
\label{sec:introduction}
Robots have well-defined actuation limits and, usually, a clear definition of the task to be completed, but the conditions of the environment in which they operate may be a source of uncertainty.
Besides environmental uncertainty, robots can also be affected by sensor noise, signal delay, and model mismatches, and these sources of error are often addressed with a feedback controller.
However, controllers have their own limitations, and their ability to execute a motion depends not only on the complexity of the trajectory but also on the control authority available to track the motion plan and counteract any external disturbances at the same time.
In general, there are two ways to improve robustness:
\begin{itemize}
    \item During \textit{control} (\cite{delprete2015addressing,plooij2015robust,xin2018modelbased}), by increasing robustness when executing a nominal motion plan.
    \item During \textit{planning} (\cite{orsolino2018application,manchester2017dirtrel}), by considering uncertainty and robot capabilities to find trajectories with larger feasibility regions that can be exploited by controllers.
\end{itemize}
Being robust at the control stage does not necessarily result in a robust execution overall if the commanded motion is not robust itself.
In fact, a bad motion plan will inherently compromise the robustness strategy of a controller.
Despite the importance of robust controllers, we believe that ensuring robustness at an earlier stage is paramount for reliable deployment of robotic systems and, for that reason, this paper tackles the problem of increasing robustness during planning.
While predicting and modeling uncertainty at the planning stage is difficult, we can exploit well-known capabilities and limitations of a system to optimize highly-robust trajectories.
We argue that, by explicitly taking into account robot-specific capabilities and computing the set of admissible forces in task-space, we can define a metric as a function of that set to find trajectories that are more capable of resisting unexpected forces.
To that end, we first propose a representation of admissible task-space forces taking into account the dynamics of the system (i.e., not limited to quasi-static scenarios).
Then, we test our hypothesis by defining an objective function based on our proposed representation, and compare it against other established objectives.
We use a direct method to formulate the optimal control problems where those objectives are employed.
This allows for straightforward definition of mathematical constraints (in the form of equalities and inequalities) on either state or control variables, as well as the computation of the force/torque capabilities of the robot as a polytope, for any of the trajectory points discretized.

The main contributions of this work are:
\begin{enumerate}
    \item Proposal of a representation of all the realizable forces given a configuration, a vector of forces/torques, and the system dynamics: the \textit{residual force polytope}.
    \item Elucidation of two models for representing force uncertainty and their combination with the residual force polytope for optimizing robust trajectories.
    \item Comparison of several objective functions from related work with an objective function based on the residual force polytope for dynamic trajectory optimization.
\end{enumerate}

\section{Related Work}
In previous work \cite{ferrolho2018whole}, we exploited the kinematic redundancy of robots with many degrees of freedom in order to select configurations more robust to torque-tracking errors.
Our method indexes a previously-sampled database efficiently, but it does not optimize the robot's ability to resist unknown external disturbances and is limited to choosing single configurations.
In contrast, this paper focus on robustness against unexpected forces and demonstrates how the states of the system can be optimized for entire trajectories to achieve more robust motions.

Other researchers have also exploited kinematic redundancy to improve robot capabilities.
For example, Yoshikawa \cite{yoshikawa1985manipulability} proposed the \emph{force manipulability ellipsoid} to take into account the ability to apply and resist forces based on the robot geometry.
Building on top of this concept, Jaquier \textit{et al.}~\cite{jaquier2018geometry} proposed a control scheme which tracks desired profiles of manipulability ellipsoids, either as the main task or as a secondary objective.
Haviland and Corke~\cite{haviland2020maximising} presented a resolved-rate motion control also making use of manipulability ellipsoids:
their real-time controller tracks the Cartesian velocity of the end-effector while maximizing the manipulability of the system.
Both \cite{jaquier2018geometry} and \cite{haviland2020maximising} employ manipulability metrics during the control stage, but such metrics can also be employed for motion planning.
An example of this is Chu \textit{et al.}'s~\cite{chu2018path} path planning algorithm for multi-arm robots:
their approach uses Yoshikawa's measure of manipulability to avoid kinematic singularities while planning complex and collision-free maneuvers.
Despite the widespread use of manipulability ellipsoids in robotics applications, the real manipulability of actuated systems is a convex polytope which cannot be represented accurately using an ellipsoid, i.e., the ellipsoid is only an approximation.
Additionally, manipulability ellipsoids make it difficult to capture and incorporate descriptions of other system constraints.
In contrast, the polytope of admissible forces that we propose in this paper is not an approximation, and allows for easy integration of extra constraints through polytope manipulation.

The ability to manipulate and intersect polytopes can be very useful.
For example, it allows the aggregation of multiple constraints into a single description of necessary conditions for feasibility of a system, provided that each individual constraint can be modeled in the form of a polytope.
For example, Audren and Kheddar \cite{audren20183d} extended 2D stability regions to 3D by accounting for possible center-of-mass accelerations in order to achieve robust multi-contact stability in whole-body posture generation.
Orsolino \textit{et al.} \cite{orsolino2018application} proposed the actuation wrench polytope and intersected it with the contact wrench cone \cite{hirukawa2006universal} to create the feasible wrench polytope.
The actuation wrench polytope is a representation of all the wrenches a robot can generate given its actuation limits.
However, it is limited to quasi-static scenarios.
In this paper, we propose a new representation that accounts for the dynamics of the system and the torques required by a nominal motion, hence, providing a description of the admissible forces for dynamic scenarios.
In \cite{orsolino2018application}, the feasibility polytope was used to optimize the center-of-mass position of a quadruped's static crawl gait.
However, due to the required computational cost, they calculated the polytope once at the beginning of the optimization and used that as a constant approximation thereafter.
As such, computing the exact polytope at every point of the trajectory during optimization and the impact of this approach on performance are two important aspects that have not been studied before, and which we address in this work.
It is also worth noting that the optimization problem in \cite{orsolino2018application} optimizes four variables in time (the positions and velocities of the center of mass in the $xy$-plane), while our problem optimizes 21 variables in time (the state and control inputs of a 7-DoF robot arm) and is therefore significantly more complex.

The idea of improving the robustness of robot motions using trajectory optimization has been explored before:
Manchester and Kuindersma \cite{manchester2017dirtrel} presented an algorithm that incorporates linear feedback, bounded disturbances, and a penalty for closed-loop deviations from a nominal trajectory.
A key advantage of their method is that the resulting control trajectories avoid \textit{bang-bang} control, and leave margins of stability for LQR feedback control around the nominal trajectory.
Our approach also retains these advantages as a result of the polytope-based objective functions and, additionally, the new representation we propose allows determining the exact margins remaining before torque saturation occurs.

The more general idea of increasing robustness of an optimization model is also important in fields outside of robotics.
For instance, Ben-Tal and Nemirovsky \cite{bental1997robust} applied the more general idea of increasing robustness of an optimization model to \gls{TTD}.
They cast \gls{TTD} problems as semidefinite programs to minimize worst-case compliance of trusses under external loads, and used additional constraints to increase their robustness.
For that, they considered not only primary loads (specified by the user), but also secondary loads from different directions and with reasonable magnitude.

\section{Preliminaries}
\label{sec:preliminaries}

\subsection{Polytopes and the Double Description Method}
\label{subsec:polytopes_and_the_double_description_method}
A convex polytope \cite{ziegler2012lectures} can be defined in one of two ways:
\begin{itemize}
    \item Vertex representation ($\mathcal{V}$-rep): a finite set of points;
    \item Half-space representation ($\mathcal{H}$-rep): a bounded intersection of a finite set of half-spaces.
\end{itemize}
For some mathematical operations, one representation has some inherent advantages over the other.
For example, the intersection of two or more polytopes is easier to perform in $\mathcal{H}$-rep than in $\mathcal{V}$-rep, and a Minkowski sum is easier to carry out in $\mathcal{V}$-rep than in $\mathcal{H}$-rep.

It may happen that a $\mathcal{V}$-rep is required when only an $\mathcal{H}$-rep is available, or vice versa---this is known as the representation conversion problem.
It is possible to convert from one representation to the other using the double-description method~\cite{fukuda1996double}.
Nonetheless, switching between representations can be computationally very expensive and should be avoided.

\subsection{Robot Model Formulation}
Consider a fully-actuated robot manipulator with $n$ degrees of freedom, a fixed base, and with an end-effector operating in an $m$-dimensional task-space.
Such a system can be parameterized with a generalized coordinates vector $\bm{q} \in \mathbb{R}^n$ and a generalized velocities vector $\bm{v} \in \mathbb{R}^n$.
The dynamics of the system are given by the equations of motion:
\begin{equation}
    \bm{M}(\bm{q})\bm{\dot{v}} + \bm{h}(\bm{q}, \bm{v}) = \bm{\tau} + \bm{J}_e^\top(\bm{q}) \bm{f}_{\mathrm{tip}},
    \label{equation:equations_of_motion}
\end{equation}
where $\bm{M}(\bm{q}) \in \mathbb{R}^{n \times n}$ is a symmetric positive-definite mass matrix,
$\bm{h}(\bm{q}, \bm{v}) \in \mathbb{R}^{n}$ is the vector of Coriolis, centrifugal, and gravity terms,
$\bm{\tau} \in \mathbb{R}^{n}$ is the vector of joint forces and torques,
$\bm{J}_e \in \mathbb{R}^{m \times n}$ is the Jacobian matrix that maps joint velocities to the linear velocity of the end-effector,
and $\bm{f}_{\mathrm{tip}} \in \mathbb{R}^m$ is a force applied to the end-effector.
The transpose of $\bm{J}_e$ maps a linear force applied at the end-effector to a vector of torques experienced at the joints of the mechanism---in the following referred to as $\bm{\tau}_{\bm{f}_{\mathrm{tip}}}$.
Conversely, we can determine an end-effector force generated from a vector of input torques with
\begin{equation}
    \bm{f}_{\mathrm{tip}} = \bm{J}^{-\top}_e \bm{\tau}_{\bm{f}_{\mathrm{tip}}}.
    \label{equation:forces_from_torques}
\end{equation}
In some cases, it may not be possible to invert $\bm{J}^{\top}_e$ because it is singular.
Likewise, kinematically redundant systems have more joints than the dimension of their task space ($n > m$) and therefore $\bm{J}^{\top}_e$ is not square and cannot be inverted.
However, for such cases, we can still solve equation~\eqref{equation:forces_from_torques} by using the Moore-Penrose pseudoinverse to invert $\bm{J}^{\top}_e$.

The mapping in equation~\eqref{equation:forces_from_torques} is instrumental for computing force polytopes, which we explain next.

\subsection{Joint Force Polytope and Force Polytope}
\label{subsec:force_polytope}
The \textit{joint force polytope}~\cite{chiacchio1997force} is an $n$-dimensional region bounded by the upper and lower actuation limits of the system.
It is described by the $2n$ bounding inequalities
\begin{equation}
    \left | \tau_{i} \right | \leq \tau_{i,\mathrm{lim}} \qquad i = 1, \cdots, n,
    \label{equation:force_polytope}
\end{equation}
where $\tau_{i,\mathrm{lim}}$ is the bound on the $i$-th joint force.

\begin{figure}[ht]
    \centering
    \includegraphics[width=0.9\linewidth]{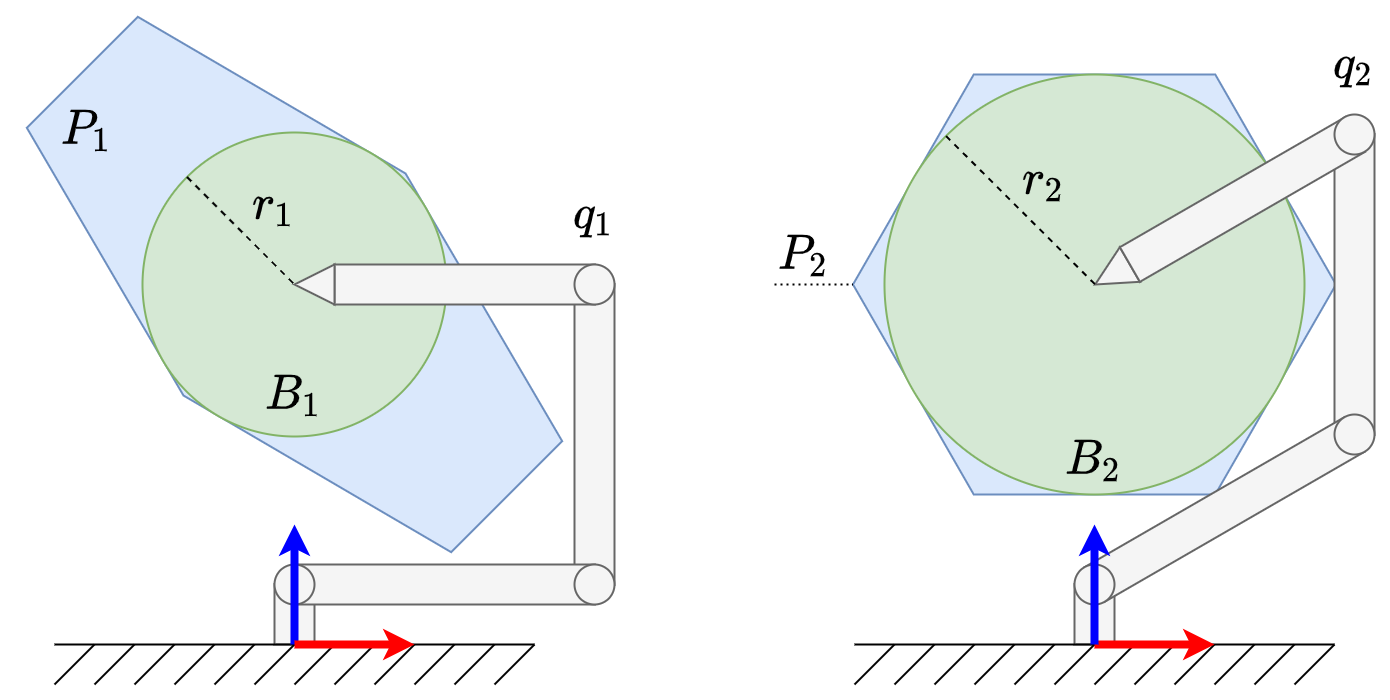}
    \caption{
        Two valid configurations for reaching the same end-effector target.
        The blue polygons $P_1$ and $P_2$ are the force polytopes of configurations $\bm{q}_1$ and $\bm{q}_2$, respectively.
        The green circles $B_1$ and $B_2$ are the largest balls centered at the end-effector that can be inscribed inside those polytopes.
        The radius of $B_1$ and $B_2$ are denoted by $r_1$ and $r_2$, and here $r_2 > r_1$.
    }
    \label{figure:diagram_redundancy}
\end{figure}

The \textit{force polytope} is the convex set of all the realizable forces by the end-effector for quasi-static scenarios, given the actuation limits of the system.
A force polytope $P_{\bm{f}}$ results from transforming a joint force polytope $P_{\bm{\tau}}$ with $P_{\bm{f}} = \bm{J}^{-\top}_e P_{\bm{\tau}}$, analogous to how equation \eqref{equation:forces_from_torques} converts a vector of joint-space forces and torques into a task-space force.
Because of this nonlinear relationship, different robot configurations result in force polytopes with different shapes.
\autoref{figure:diagram_redundancy} illustrates this trait: two redundant configurations, $\bm{q}_1$ and $\bm{q}_2$, reach the same end-effector target, but their respective force polytopes, $P_1$ and $P_2$, have distinct shapes.

\section{Residual Force Polytope}
In \autoref{subsec:force_polytope} we have reviewed what a force polytope is and how it results from the mapping of the actuation limits of a robot into the task-space.
The force polytope is limited to quasi-static scenarios and, besides the kinematic configuration of the robot, it does not take into account any information about the task being performed.

We propose a new representation called the \textit{residual force polytope}, which takes the dynamics of the robot into account, as well as the nominal forces and torques required by a task.
We define the residual forces and torques of a robot state as the difference between the absolute actuation limits and a given vector of joint forces and torques.
Residual forces and torques are important to deal with disturbances, as they represent the control authority left in a system after accounting for the task at hand.
The residual force polytope is the result of transforming those residual forces and torques with $\bm{J}^{-\top}_e$, similarly to equation \eqref{equation:forces_from_torques}.
In summary, the residual force polytope is a subset of its force polytope counterpart.
It represents exclusively the forces that the robot is capable of resisting (as a secondary task) while tracking a nominal trajectory as its primary task.

\begin{figure}[ht]
    \centering
    \includegraphics[width=0.9\linewidth]{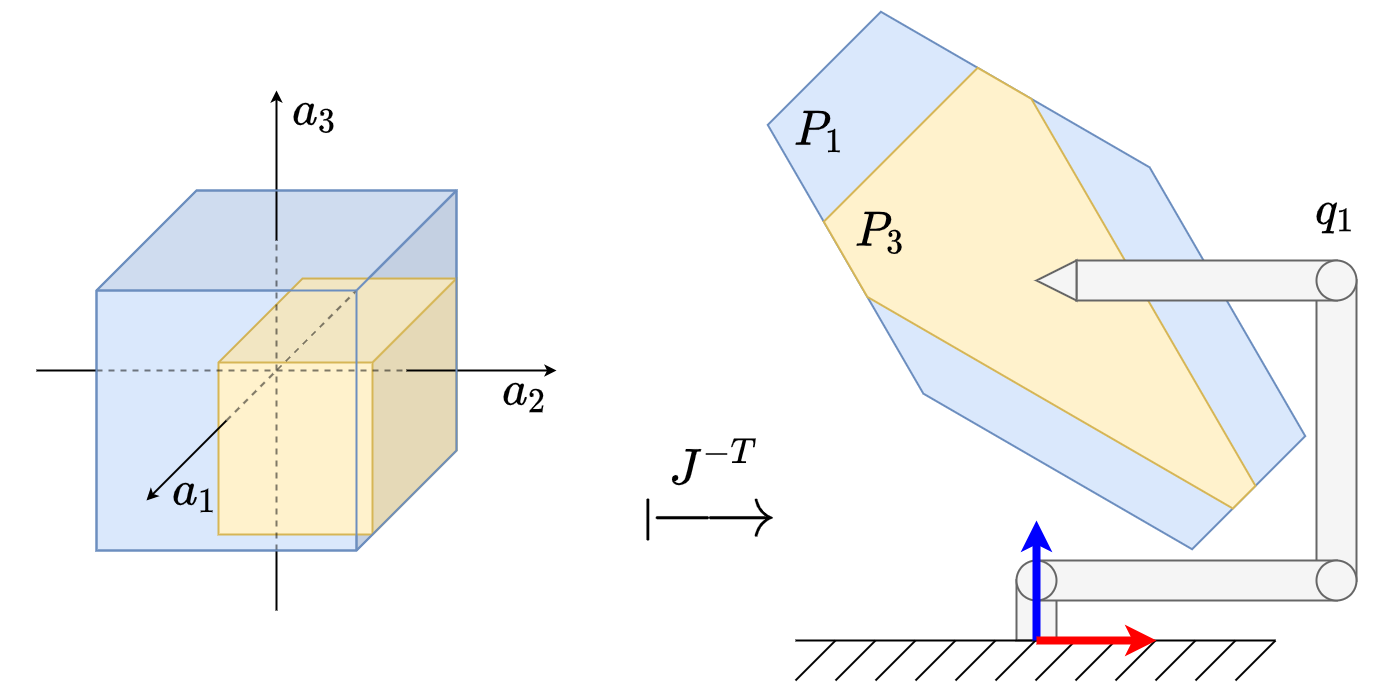}
    \caption{
        Equation \eqref{equation:forces_from_torques} transforms actuation-space representations (on the left) into task-space representations (on the right).
        The blue polyhedron on the left is the \textit{joint force polytope}, and by taking into account a given vector of torques it is reduced along some dimensions into the yellow polyhedron.
        The yellow polygon $P_3$ (on the right) is the \textit{residual force polytope}.
    }
    \label{figure:diagram_residual}
\end{figure}

\autoref{figure:diagram_residual} shows the relationship between forces/torques in actuation-space and forces in task-space.
For convenience of illustration, it displays a planar manipulator with three degrees of freedom in joint-space and two-dimensional task-space forces.
The figure highlights how the residual force polytope $P_3$ is obtained for a given configuration $\bm{q}_1$.

\section{Modeling Force Uncertainty}
As we have seen so far, polytopes are useful to represent and model regions of interest in space.
But in addition to this, we may want to extract a single metric that quantifies one of those regions.
For example, given a configuration $\bm{q}_1$ and its corresponding force polytope $P_1$, we may want to know how robust that configuration is with respect to forces applied at the end-effector of the robot.

\subsection{Largest Ball Inscribed in a Polytope}
One way to tackle this problem is to consider the worst-case scenario, i.e., the situation with most uncertainty, where a force could originate from any given direction.
In order to represent this uncertainty, we can use a ball to model a set of forces with any given direction and with a magnitude ranging from \SI{0}{\newton} to the radius of the ball.
Then, if we constrain the ball to be centered at the end-effector, and maximize the size of the ball without exceeding the boundaries of the force polytope, we obtain the set of all forces that the robot is able to deliver without saturating its torque limits.
Consequently, the radius of this ball denotes the magnitude of the greatest force that the robot can counteract, and it can be used as a metric for isotropic robustness of a configuration.\footnote{%
    In light of directional uncertainty, an isotropic robustness metric is more useful than other general quantities like the overall volume of a polytope.
}
For example, both $\bm{q}_1$ and $\bm{q}_2$ shown in \autoref{figure:diagram_redundancy} solve the same reaching task, but $\bm{q}_2$ is more robust than $\bm{q}_1$ because $r_2 > r_1$.

The center of the largest ball $B$ inscribed in a bounded set of non-empty interior is known as the \textit{Chebyshev center}~\cite{boyd2004convex}.
We can find the Chebyshev center of a polytope $P$ by solving a \gls{LP} problem
where the center of the ball~$B$ and its radius~$r$ are the decision variables,
and the goal is to maximize $r$ subject to the constraint $B \subseteq P$.
In our work, we are interested in a similar problem but where the center of the ball lies at the origin of the end-effector frame.
This is because we only care about the forces that can be applied specifically to the end-effector.
Therefore, we formulate an \gls{LP} problem which maximizes $r$ subject to the constraint $B \subseteq P$, but where the only decision variable is the radius $r$ (since the center of $B$ is known and given by the forward kinematics function of the robot's current configuration).

\begin{figure}[t]
    \centering
    \subfigure[For legged robots locomoting on complex terrains, the direction of the terrain normals can change greatly with small variations in the contact location, leading to very different contact forces applied to the feet.]{\includegraphics[width=0.48\linewidth]{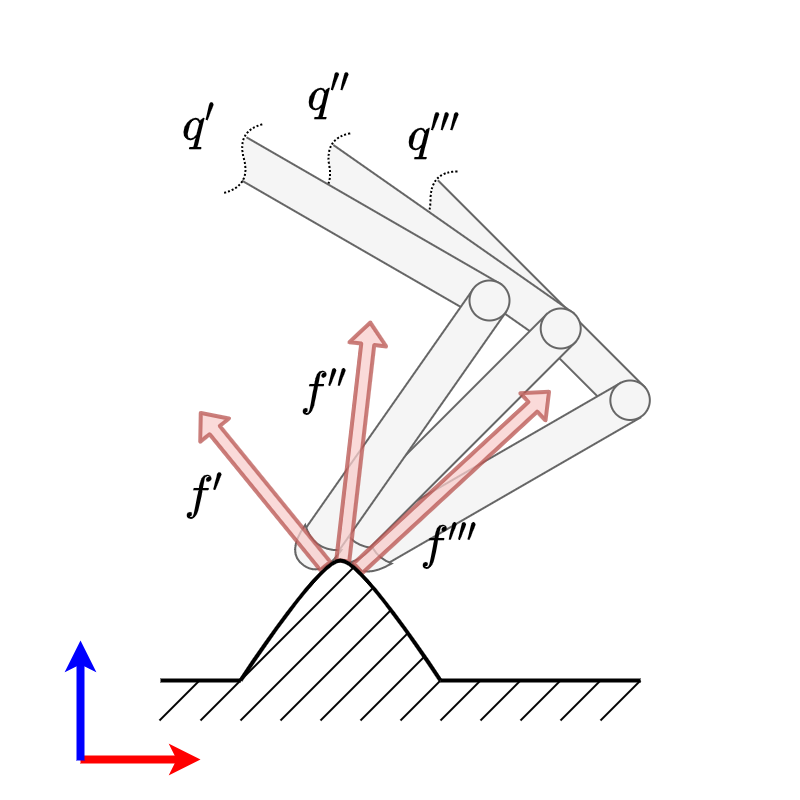}}\hfill%
    \subfigure[In the context of a manipulation task, the direction of action/reaction forces for lifting a box attached to a rope is well-known \textit{a priori}, and regardless of the weight of the box.]{\includegraphics[width=0.48\linewidth]{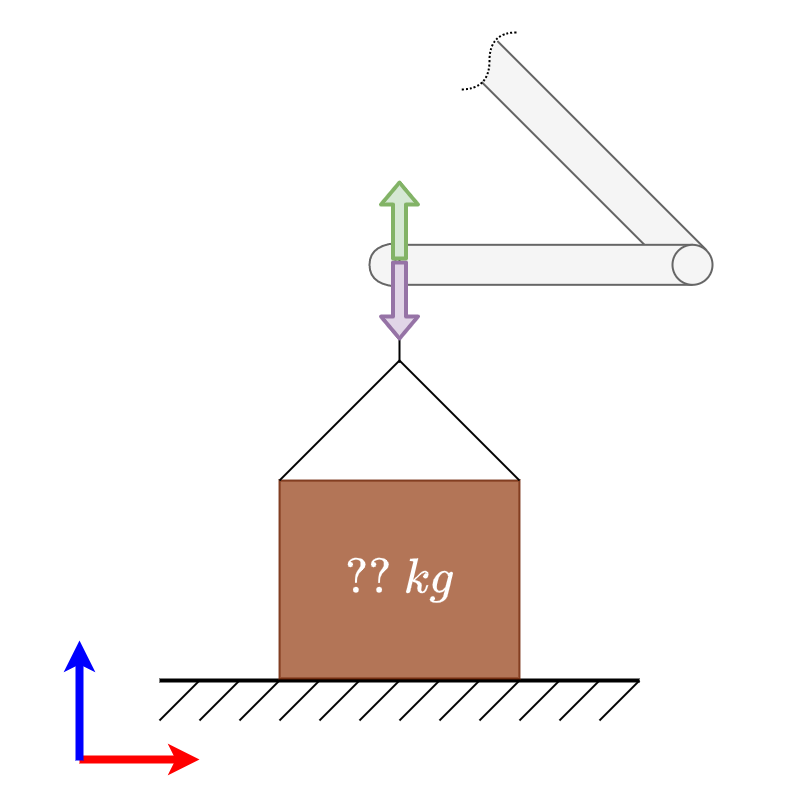}}\hfill%
    \caption{
        Illustration of two different levels of uncertainty concerning the direction of interaction forces for two real-world scenarios.
        On the left, factors such as controller tracking errors or noisy state estimation can ultimately lead to inaccurate foot placement, which in turn, and depending on the terrain, can induce forces applied in unexpected directions (high direction uncertainty).
        In contrast, on the right, the forces are expected to be close-to-vertical due to the nature of the task (low direction uncertainty).
    }\label{figure:diagram_leg_and_box}
\end{figure}

\subsection{Largest Intersection with a Polytope}
\label{subsec:largest_intersection_with_a_polytope}
\begin{figure}[ht]
    \centering
    \includegraphics[width=\linewidth]{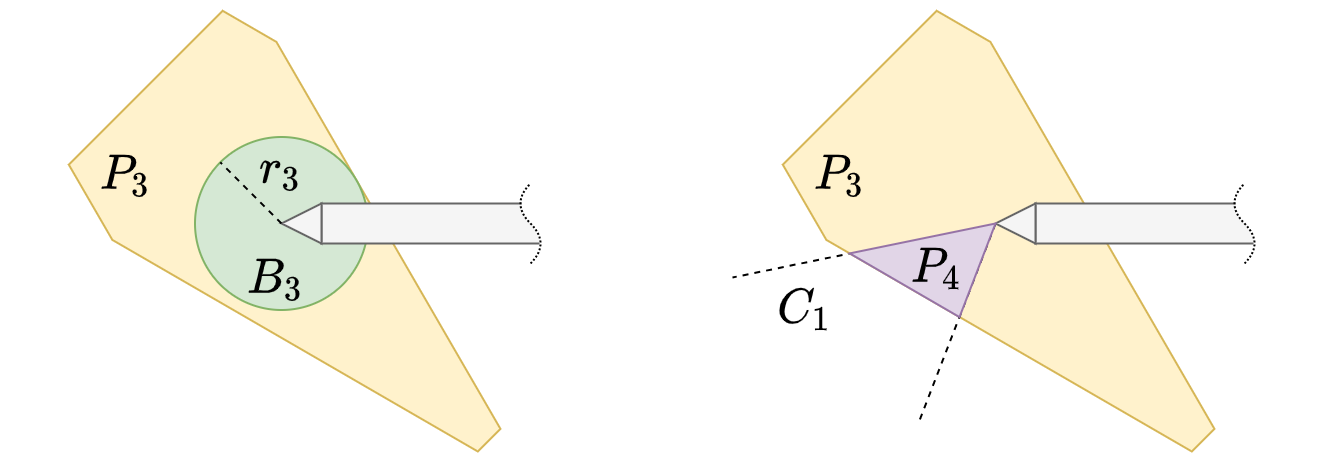}
    \caption{
        Two distinct models for representing force disturbances:
        the ball $B_3$ models unexpected forces, whereas the cone $C_1$ models the direction of an expected force.
        The aperture of the cone is proportional to the uncertainty of the force direction.
        The purple polygon $P_4$ results from the intersection of the \textit{residual force polytope} with the cone, i.e., $P_4 = P_3 \cap C_1$.
    }\label{figure:diagram_direction}
\end{figure}

The previous subsection demonstrated how to calculate the robustness of a robot to completely unknown external disturbances.
However, there are cases where the interaction between the robot and its environment is not fully uncertain.
As an example, consider a task where the robot needs to open or close a door of unknown mass:
the robot may not know \textit{a priori} how much force is needed to solve the task, but the door can only open or close in a specific way---see \autoref{figure:diagram_leg_and_box} for a further example.
The direction of the interaction can therefore be exploited to our advantage.
We can use a cone to model the set of forces originating from some expected direction and applied to the end-effector:
the cone axis is aligned with the expected direction, the cone apex is fixed at the end-effector frame, and the aperture of the cone represents the prediction uncertainty of the force direction.
Then, if we intersect the cone with a force polytope, we obtain a subset of the forces in the cone which the robot can cancel out within its actuation limits.
Consequently, the volume of the resulting intersection is proportional to how much the robot is capable of resisting forces modeled by the cone, and it can be used as a surrogate metric of robustness to expected forces.
An example of modeling expected forces using this approach is illustrated in \autoref{figure:diagram_direction}, where the intersection of a cone $C_1$ with a residual force polytope $P_3$ results in the purple polygon $P_4$, i.e., $P_4 = P_3 \cap C_1$.

In this section, we showed that redundant configurations result in different capabilities to counteract external forces applied to the end-effector.
We proposed a representation for modeling those capabilities, and discussed two robustness metrics that can be extracted from it.
In the next section, we will demonstrate how to formulate a trajectory optimization problem with objective functions that employ those metrics in order to plan dynamic motions more robust to unexpected forces through exploitation of kinematic redundancy.

\section{Optimization of Robust Trajectories}
Trajectory optimization is a process that allows to compute control trajectories as functions of time that drive a system from an initial state towards a final state while satisfying a given set of constraints \cite{betts2010practical}.
In robotics, the problem is a second-order dynamical system governed by the equations of motion \eqref{equation:equations_of_motion}.

\emph{Direct transcription}~\cite{vonstryk1992direct} is a popular approach within trajectory optimization and works by transcribing a continuous problem into a constrained nonlinear optimization problem by means of explicit discretization of the state and control trajectories.
The result of this transcription is the formulation of a large and sparse nonlinear problem which can be solved using a large-scale nonlinear programming solver~\cite{betts2010practical}.

We have chosen direct transcription to demonstrate how the residual force polytope can be used to plan robust and dynamic trajectories.
Thanks to the discretization of states and controls, the configuration of the robot and the commanded torques are represented as decision variables for every discrete point of the trajectory.
This means that all the ``ingredients'' required to compute the polytope representations (discussed in previous sections) are readily available as decision variables.
Similarly, it also means that it is easy to define equality and inequality constraints using those decision variables, which general off-the-shelf \gls{NLP} solvers can then handle during problem resolution.
In contrast, the most popular alternative, \gls{DDP}~\cite{mayne1966second}, does not allow for easy definition of constraints (neither equalities nor inequalities).
There are variations and extensions to classical \gls{DDP} which attempt to mitigate this inconvenience (e.g., \cite{tassa2014control,giftthaler2018family,mastalli2020crocoddyl}), but this topic is still a subject of ongoing research and those variations are not yet mature enough.

In summary, we chose direct transcription because:
\begin{itemize}
    \item Discretization of both states and controls is particularly convenient for computing polytope representations;
    \item Defining general state and path constraints using direct transcription is more straightforward than alternatives;
    \item Its simplicity of formulation and implementation.
\end{itemize}

\subsection{Problem Formulation}
We divide the trajectory into $N$ equally spaced segments
\begin{equation}
    t_I = t_1 < t_2 < \dots < t_M = t_F,
\end{equation}
where $t_I$ and $t_F$ are the start and final instants, respectively.
Thus, the number of discretized \textit{mesh points} is $M = N + 1$.
Let $x_k \equiv x(t_k)$ and $u_k \equiv u(t_k)$ be the values of the state and control variables at the $k$-th mesh point, respectively.
We treat $x_k \triangleq \{ \bm{q}_k, \bm{v}_k \}$ and $u_k \triangleq \{ \bm{\tau}_k \}$ as a set of nonlinear programming variables, and formulate the trajectory optimization problem as:
\begin{equation}
    \begin{aligned}
        \argmin_{\bm{\xi}} \qquad   & \sum_{k=1}^{M}\ g(x_k, u_k) \\
        \mathrm{subject\ to} \qquad & \dot{x} = f(x, u)           \\
                                    & x_k \in \mathcal{X}         \\
                                    & u_k \in \mathcal{U}
    \end{aligned}
    \label{equation:nlp}
\end{equation}
where $\bm{\xi}$ is the vector of decision variables,
$g(\cdot, \cdot)$ is a cost function,
$\dot{x} =f(x, u)$ gives the nonlinear dynamics of the system,
and $\mathcal{X}$ and $\mathcal{U}$ are sets of feasible states and control inputs enforced by a set of equality and inequality constraints.
The vector of decision variables $\bm{\xi}$ results from aggregating the generalized coordinates, generalized velocities, and control inputs of every mesh point:
\begin{align}
    \bm{\xi} \triangleq \{ \bm{q}_1, \bm{v}_1, \bm{\tau}_1, \cdots, \bm{q}_{N}, \bm{v}_{N}, \bm{\tau}_{N}, \bm{q}_M, \bm{v}_M \}.\footnotemark
\end{align}
\footnotetext{The control inputs at the final state $\bm{\tau}_M$ need not be discretized.}%

\subsection{Constraints}
We want to optimize trajectories that are consistent with the full dynamics of the robot, do not exceed the kinematic and actuation limits of the robot, and use the end-effector for a given task.
We formulate all these requirements as equality and inequality constraints which the solver must respect.

\subsubsection{End-effector Task}
The exemplar task we use for this evaluation is to move the end-effector of a multi-DoF robot arm from an initial point $\bm{p}_I$ to a final point $\bm{p}_F$:
\begin{equation}
    f_\mathrm{FK}(\bm{q}_1) = \bm{p}_I
    \quad\mathrm{and}\quad
    f_\mathrm{FK}(\bm{q}_M) = \bm{p}_F
\end{equation}
where $f_\mathrm{FK}(\cdot)$ is the forward kinematics function.
In addition, the end-effector must always lie on a rectangular surface $R$ positioned in its workspace:
\begin{equation}
    f_\mathrm{dist}(R, f_\mathrm{FK}(\bm{q}_k)) = 0 \qquad \forall k = 1 : M
\end{equation}
where $f_\mathrm{dist}(\cdot)$ is the distance between a surface and a point.
This task is analogous to drawing a line on a whiteboard using a marker attached to the end-effector, where the initial and final points are given and the path taken by the end-effector does not matter as long as it does not lift the tip of the marker off from the surface of the whiteboard.

\subsubsection{System Dynamics}
We enforce the nonlinear dynamics of the system with a finite set of \textit{defect constraints}.
In summary, defect constraints are nonlinear equality constraints that ensure consistency between two consecutive mesh points.\footnote{See Chapter 3.4 of Betts~\cite{betts2010practical} for further detail regarding defect constraints.}
They make sure that the robot state at the next time step ($x_{k+1}$) matches the propagation of the previous robot state ($x_k$) given its control inputs ($u_k$).
In our formulation, we define these constraints as
\begin{equation}
    x_{k+1} - \big( x_k + h \cdot f(x_k, u_k) \big) = 0.
\end{equation}
For simplicity of exposition, we integrate the differential equations of the system dynamics using the explicit Euler method, where $h = (t_F - t_I) / N$ is the integration time step.

\subsubsection{Initial and Final Joint Velocities}
We enforce the initial and final velocities of every joint to be zero with
\begin{align}
    \bm{v}_1 = \bm{v}_M = \bm{0}.
\end{align}

\subsubsection{Bounds of the Decision Variables}
We constrain the joint positions, velocities, and torques to be within their corresponding lower and upper bounds:
\begin{align}
     & \bm{q}_{\mathrm{lb}}    \le \bm{q}_k    \le \bm{q}_{\mathrm{ub}}    &  & \forall k = 1 : M     \\
     & \bm{v}_{\mathrm{lb}}    \le \bm{v}_k    \le \bm{v}_{\mathrm{ub}}    &  & \forall k = 1 : M     \\
     & \bm{\tau}_{\mathrm{lb}} \le \bm{\tau}_k \le \bm{\tau}_{\mathrm{ub}} &  & \forall k = 1 : M - 1
\end{align}

\begin{figure*}[t]
    \centering
    \subfigure[Objective $g_A$]{\includegraphics[width=0.2\linewidth]{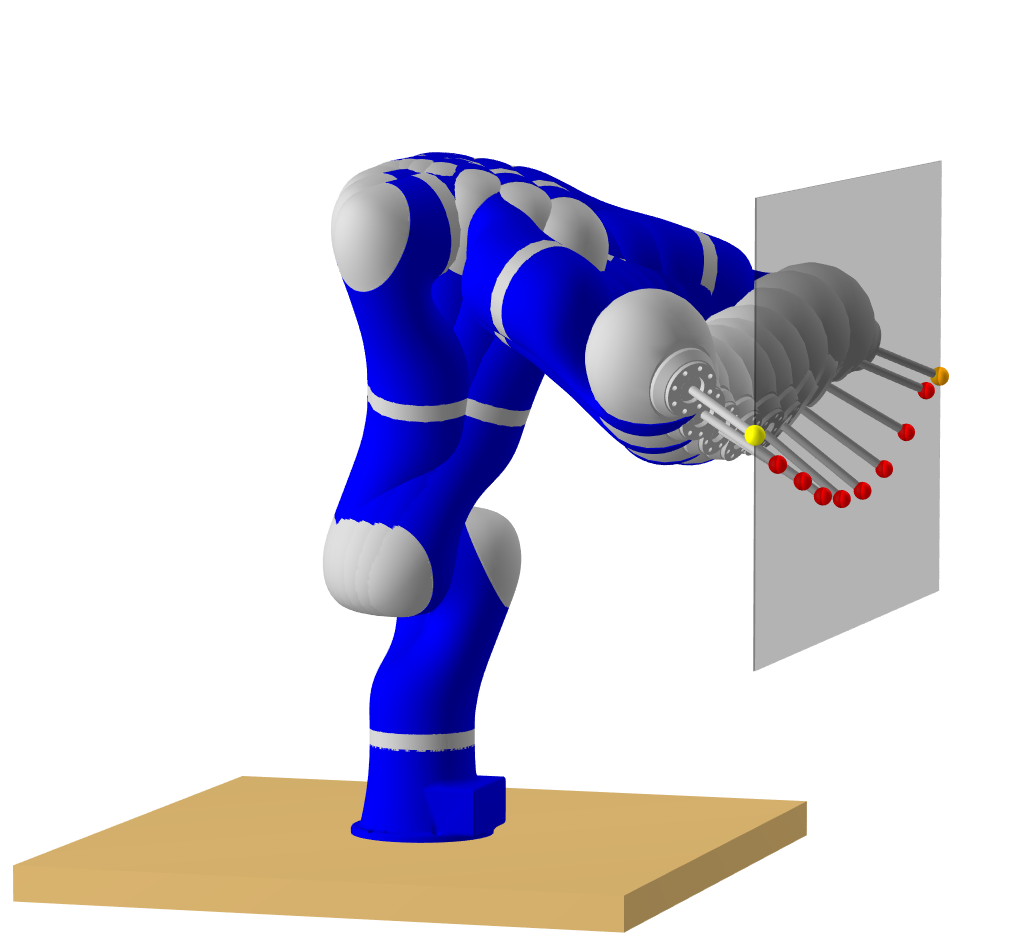}}%
    \subfigure[Objective $g_B$]{\includegraphics[width=0.2\linewidth]{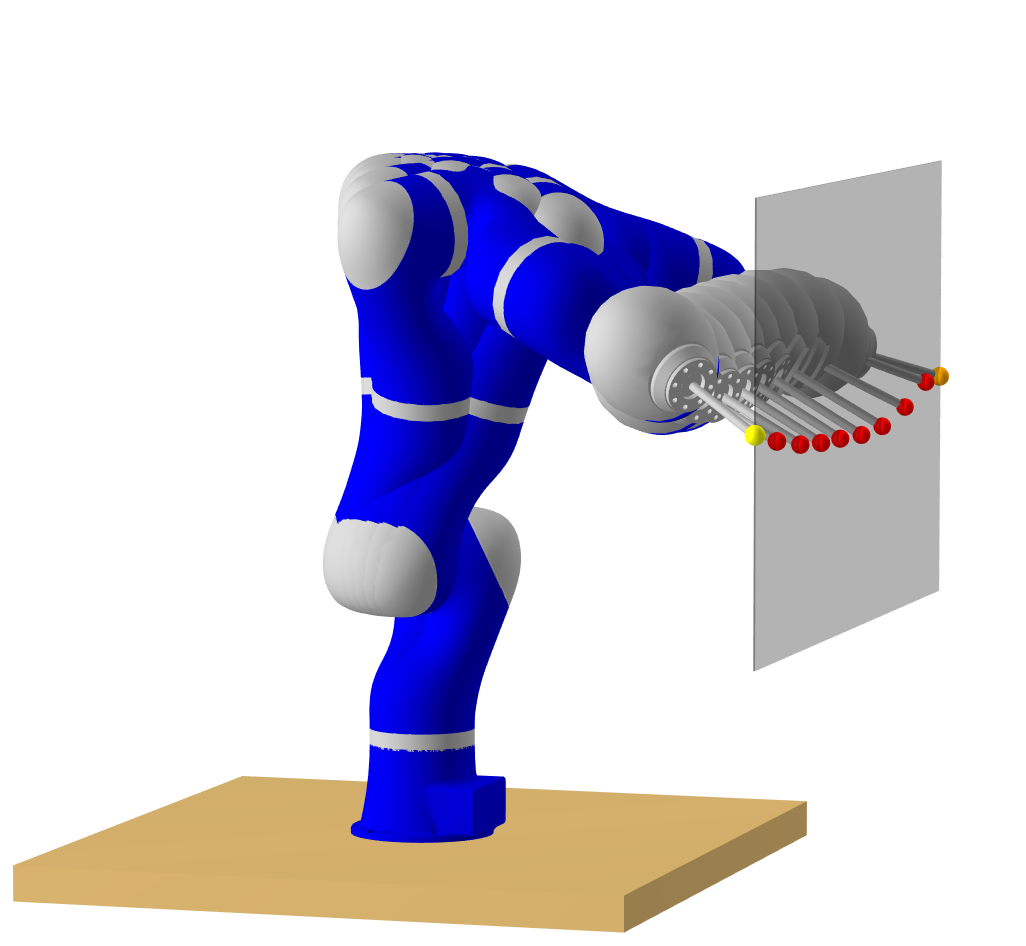}}%
    \subfigure[Objective $g_C$]{\includegraphics[width=0.2\linewidth]{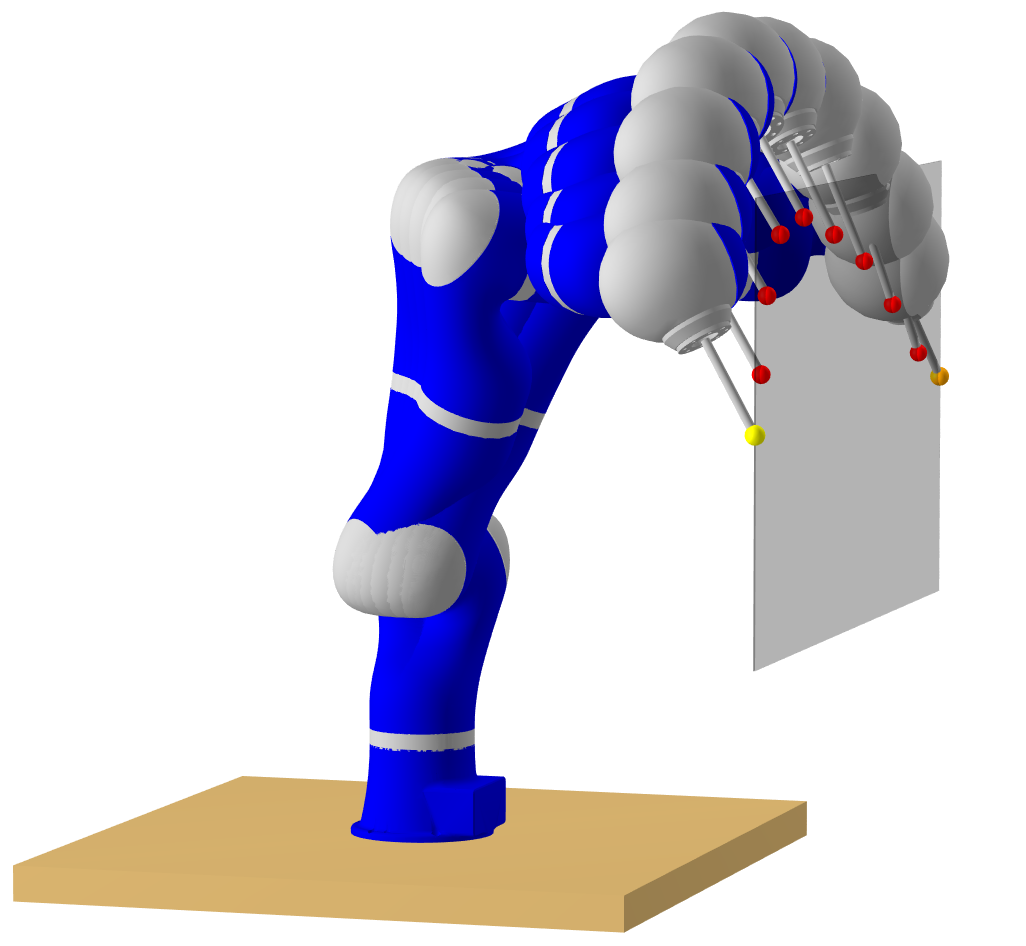}}%
    \subfigure[Objective $g_D$]{\includegraphics[width=0.2\linewidth]{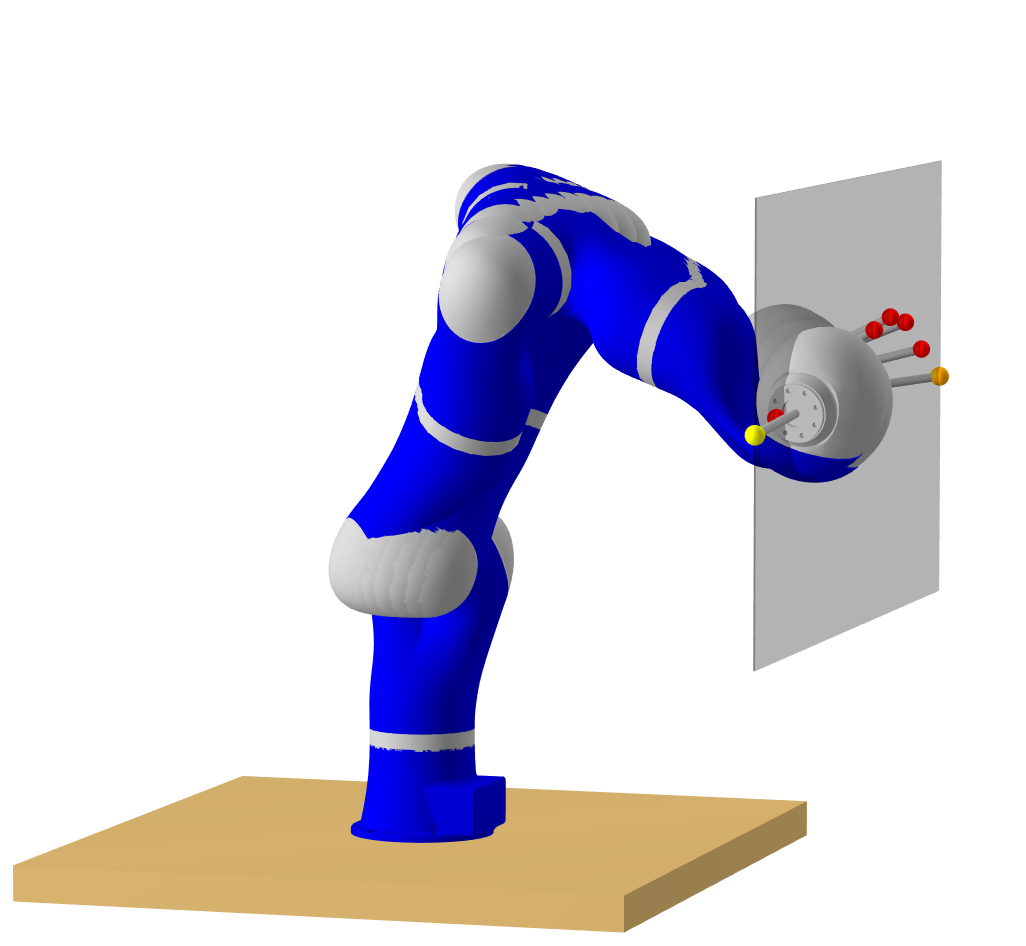}}%
    \subfigure[Objective $g_E$]{\includegraphics[width=0.2\linewidth]{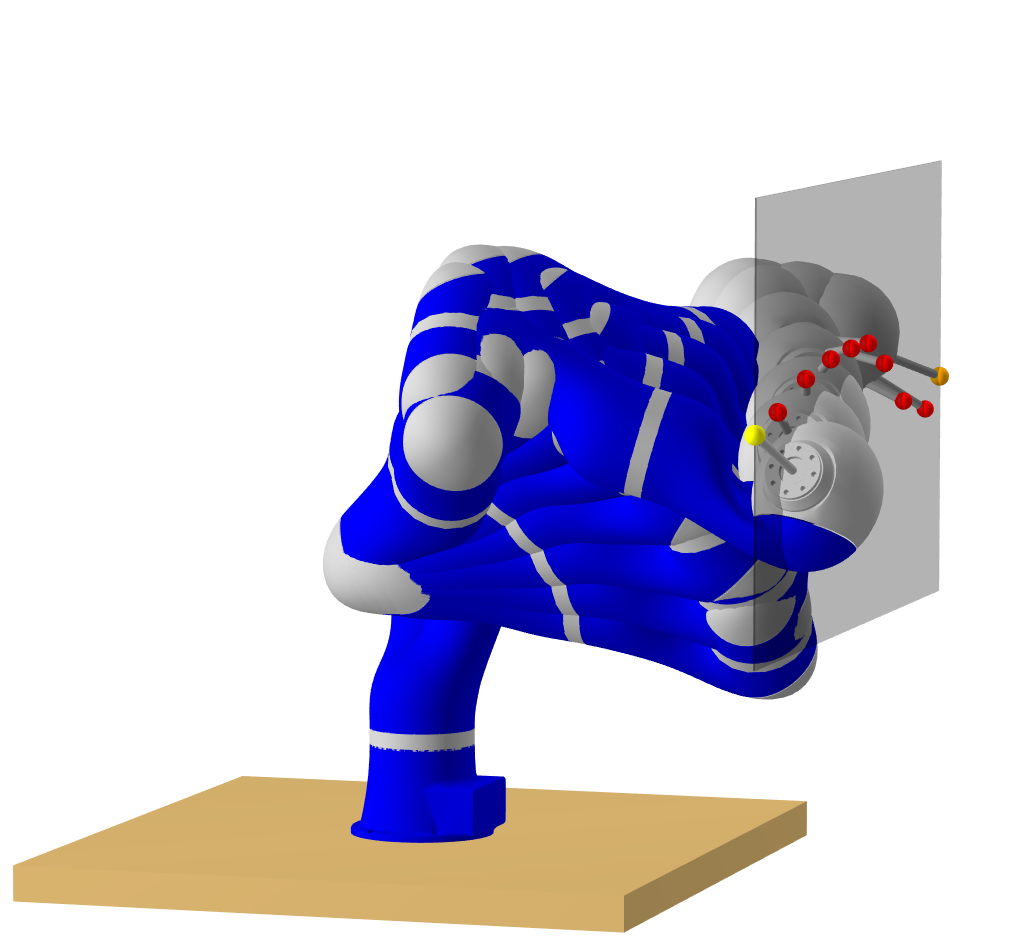}}\\%
    \subfigure[Objective $g_F$, \SI{  0}{\degree}]{\includegraphics[width=0.2\linewidth]{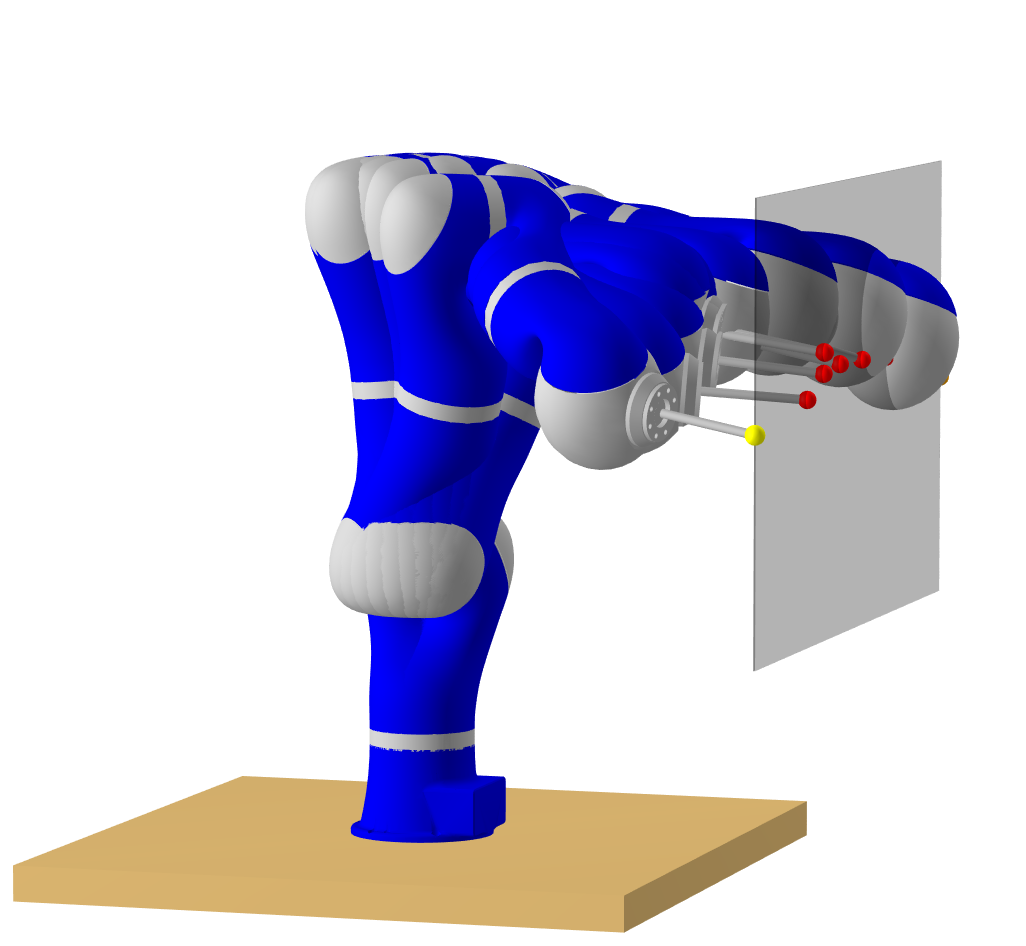}}%
    \subfigure[Objective $g_F$, \SI{ 90}{\degree}]{\includegraphics[width=0.2\linewidth]{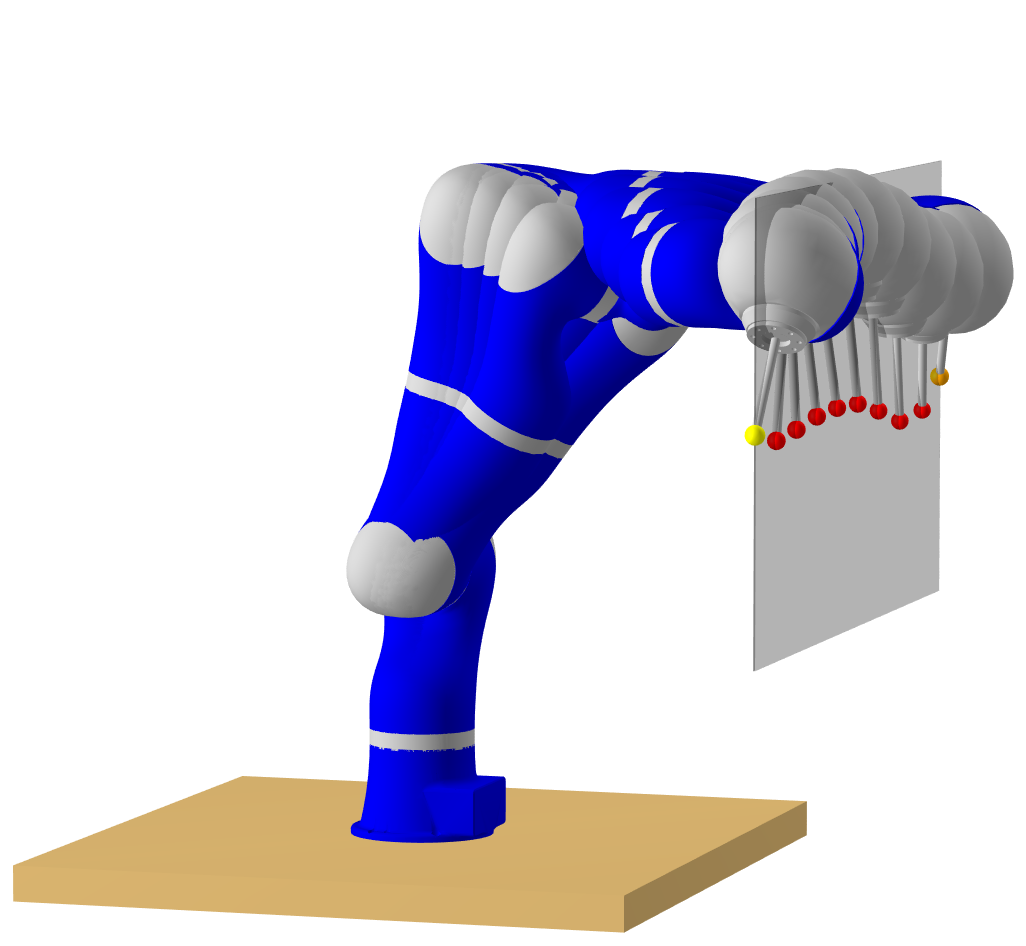}}%
    \subfigure[Objective $g_F$, \SI{180}{\degree}]{\includegraphics[width=0.2\linewidth]{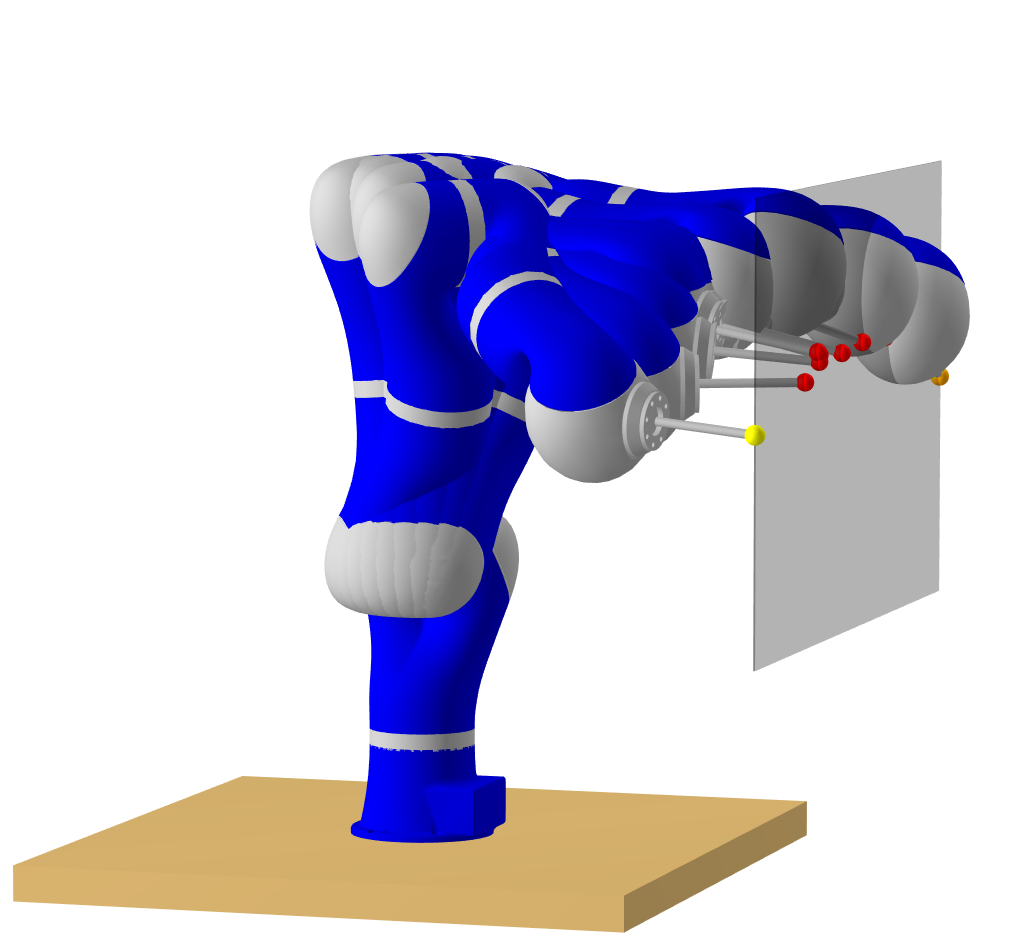}}%
    \subfigure[Objective $g_F$, \SI{270}{\degree}]{\includegraphics[width=0.2\linewidth]{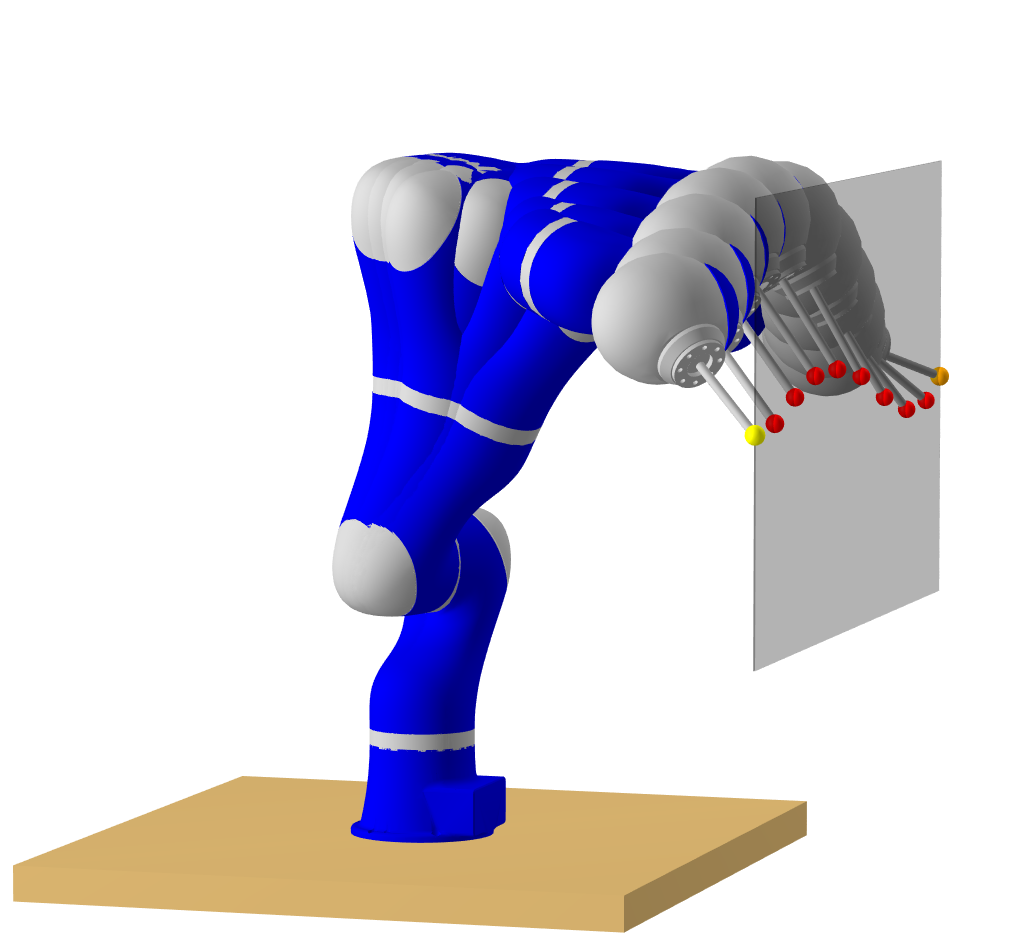}}%
    \caption{
        Visualization of the trajectories obtained using the interior-point method without a payload.
        The configuration samples are equally spaced in time.
        The orange and yellow spheres denote the start and final targets for the end-effector.
        Trajectories generated with $g_F$ depend on a specific direction; here we show four examples: \SI{0}{\degree}, \SI{90}{\degree}, \SI{180}{\degree}, and \SI{270}{\degree}.
        These angles correspond to being robust to forces originating from the front, left, back, and right sides of the robot.
    }\label{figure:trajectories}
\end{figure*}

\subsection{Objectives}
\label{subsec:objectives}
There are many objective functions which could be used to achieve different optimal results under the same problem constraints.
We will now list some well-known objectives as well as our own.
Later, in our experiments, we will compare the obtained trajectories against each other in terms of their robustness, torque expenditure, and computation time.

It is typical in optimal control to use energy as a cost, and this is usually formulated as a minimization of torques:
\begin{equation}
    g_A : \quad \min_{\bm{\xi}} \quad \sum_{k=1}^{M} \bm{\tau}_k^\top \bm{\tau}_k
    \label{equation:objective_a}
\end{equation}

In order to avoid torque saturation, we can define a simple objective function to maximize residual actuator torques:
\begin{equation}
    g_B : \quad \max_{\bm{\xi}} \quad \sum_{k=1}^{M} \left ( \bm{\tau}_\mathrm{lim} - \bm{\tau} \right )^\top \left ( \bm{\tau}_\mathrm{lim} - \bm{\tau} \right )
    \label{equation:objective_b}
\end{equation}

Yoshikawa~\cite{yoshikawa1985manipulability} defined a quantitative measure of manipulability as
$w = \sqrt{ \mathrm{det} \left( \bm{J}_e \bm{J}_e^\top \right) }.$
Later, Chiacchio \textit{et al.} \cite{chiacchio1997force} proposed a more accurate definition by scaling the joint forces with
$\bm{W} = \mathrm{diag}\left(1/\tau_{1,\mathrm{lim}}, \cdots, 1/\tau_{n,\mathrm{lim}}\right)$, which allowed to define a scaled Jacobian
$\bm{J'}^\top_e = \bm{W} \bm{J}^\top_e$
and a more accurate measure of manipulability
$w' = \sqrt{ \mathrm{det} \left ( \bm{J'}_e \bm{J'}^\top_e \right ) }$.
For our formulation, we can maximize the manipulability of every configuration in a discretized trajectory with the following objective:
\begin{equation}
    g_C : \quad \max_{\bm{\xi}} \quad \sum_{k=1}^{M} w'_t
    \label{equation:objective_c}
\end{equation}

We can also define objectives with metrics extracted from polytopes.
Let us denote the force polytope of a configuration as $P_k \equiv P(\bm{q}_k)$.
Similarly to \cite{orsolino2018application}, and assuming static equilibrium, we can maximize the robustness to external forces from any given direction with:
\begin{equation}
    g_D : \quad \max_{\bm{\xi}} \quad \sum_{k=1}^{M} B_{r}(P_k)
    \label{equation:objective_d}
\end{equation}
where $B_{r}(\cdot)$ denotes the radius of the largest ball centered at the end-effector and inscribed in the given polytope.

For the dynamic scenario, let us consider the residual force polytope as $P'_k \equiv P'(\bm{q}_k, \bm{\tau}_k)$, which is the novel representation we propose in this paper.
Analogous to  \eqref{equation:objective_d}, we can maximize the largest ball centered at the end-effector and inscribed in $P'_k$ for every mesh point with:
\begin{equation}
    g_E : \quad \max_{\bm{\xi}} \quad \sum_{k=1}^{M} B_{r}(P'_k)
    \label{equation:objective_e}
\end{equation}

The last objective function we consider in this work is the intersection of the residual force polytope with a cone that models an expected force but with some level of uncertainty---we proposed this in \autoref{subsec:largest_intersection_with_a_polytope}.
An objective function that maximizes the robustness in this scenario is:
\begin{equation}
    g_F : \quad \max_{\bm{\xi}} \quad \sum_{k=1}^{M} P_{\mathrm{vol}}(P'_k \cap C_k)
    \label{equation:objective_f}
\end{equation}
where $P_{\mathrm{vol}}(\cdot)$ denotes the volume of a given polytope, and $C_k \equiv C(t_k)$ is a cone modeling a disturbance at instant $t_k$.

\begin{figure*}[t]
    \centering
    \subfigure{\includegraphics[width=\linewidth]{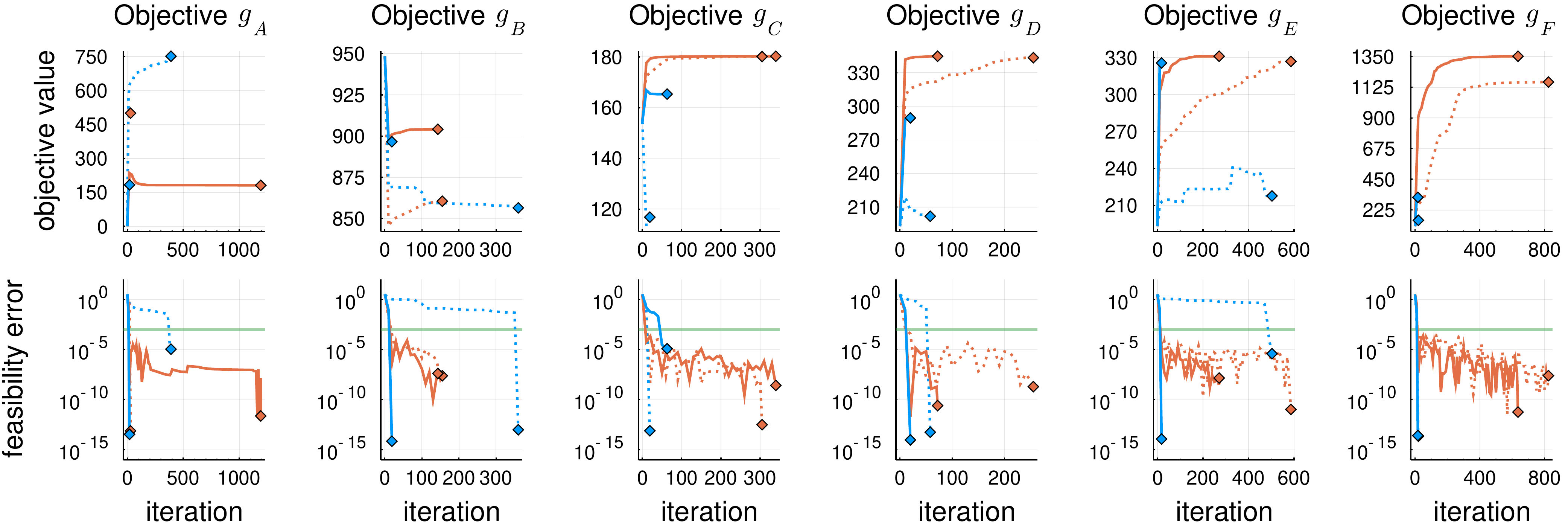}}\hfill%
    \subfigure{\includegraphics[width=0.45\linewidth]{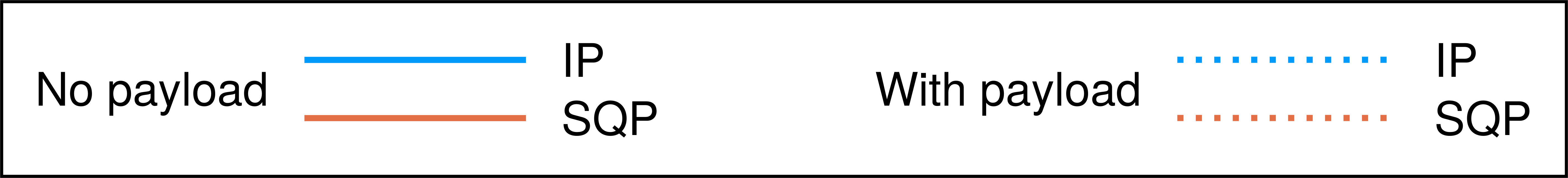}}\hfill%
    \caption{
        These plots show, for each function $g_A$--$g_F$, the evolution of the objective value and the feasibility error along the solver iterations.
        In the feasibility plots, the faint-green line at $y = 10^{-3}$ denotes the absolute tolerance under which a problem is considered feasible.
        We can see that all metrics were able to handle the payload.
        We can also see the back-and-forth progression of feasibility error for the SQP method due to excessive pivoting.
    }\label{figure:iter_obj_feas}
\end{figure*}

\begin{table*}[t]
    \small
    \centering
    \caption{Convergence times in seconds. Mean and standard deviation were calculated from 4 samples.}
    \label{table:times_convergence}
    \begin{tabular}{
        cc
        S[table-format=2.2]@{\,\( \pm \)\,}S[table-format=1.2]
        S[table-format=2.2]@{\,\( \pm \)\,}S[table-format=1.2]
        S[table-format=2.2]@{\,\( \pm \)\,}S[table-format=1.2]
        S[table-format=4.2]@{\,\( \pm \)\,}S[table-format=2.2]
        S[table-format=4.2]@{\,\( \pm \)\,}S[table-format=3.2]
        S[table-format=5.2]@{\,\( \pm \)\,}S[table-format=3.2]
        }
        \toprule
                                                 &                &
        \multicolumn{2}{c}{$g_A$ (\si{\second})} &
        \multicolumn{2}{c}{$g_B$ (\si{\second})} &
        \multicolumn{2}{c}{$g_C$ (\si{\second})} &
        \multicolumn{2}{c}{$g_D$ (\si{\second})} &
        \multicolumn{2}{c}{$g_E$ (\si{\second})} &
        \multicolumn{2}{c}{$g_F$ (\si{\second})}                                                                                                                        \\
        \midrule
        \multirow{2}{*}{No payload}              & Interior Point & 0.07  & 0.01 & 0.08  & 0.01 & 0.11  & 0.01 & 150.46  & 0.28  & 261.66  & 3.42   & 384.28   & 0.61   \\
                                                 & Active Set     & 64.50 & 0.85 & 22.51 & 0.31 & 18.21 & 0.21 & 590.44  & 5.27  & 3698.82 & 9.99   & 14354.95 & 88.38  \\
        \midrule
        \multirow{2}{*}{With payload}            & Interior Point & 0.82  & 0.02 & 0.24  & 0.01 & 0.11  & 0.01 & 1529.20 & 2.97  & 1120.90 & 1.71   & 512.91   & 1.10   \\
                                                 & Active Set     & 14.44 & 0.32 & 25.87 & 0.47 & 20.99 & 0.54 & 2177.93 & 40.91 & 8195.97 & 184.67 & 19902.78 & 369.43 \\
        \bottomrule
    \end{tabular}
\end{table*}

\begin{table*}[t]
    \small
    \centering
    \caption{Number of function evaluations and gradient evaluations of the problem constraints.}
    \label{table:ip_sqp_constraints_fevals_gevals}
    \begin{tabular}{ccrrrrrrrrrrrrrr}
        \toprule
                                                           &                &   &
        \multicolumn{6}{c}{Number of function evaluations} &                &
        \multicolumn{6}{c}{Number of gradient evaluations}                                                                                                                         \\
                                                           &                &   & $g_A$ & $g_B$ & $g_C$ & $g_D$ & $g_E$ & $g_F$ &  & $g_A$ & $g_B$ & $g_C$ & $g_D$ & $g_E$ & $g_F$ \\
        \midrule
        \multirow{2}{*}{No payload}                        & Interior Point &   & 509   & 484   & 1734  & 571   & 952   & 953   &  & 21    & 20    & 64    & 21    & 19    & 19    \\
                                                           & Active Set     &   & 34629 & 4084  & 11066 & 2285  & 15041 & 34909 &  & 1192  & 144   & 341   & 73    & 272   & 637   \\
        \midrule
        \multirow{2}{*}{With payload}                      & Interior Point &   & 9399  & 8686  & 542   & 1594  & 25217 & 1104  &  & 391   & 361   & 20    & 59    & 504   & 22    \\
                                                           & Active Set     &   & 787   & 4829  & 9992  & 8470  & 32655 & 45864 &  & 30    & 156   & 306   & 256   & 587   & 825   \\
        \bottomrule
    \end{tabular}
\end{table*}

\begin{figure*}[t]
    \centering
    \subfigure[Interior-Point method.]{\includegraphics[width=0.48\linewidth]{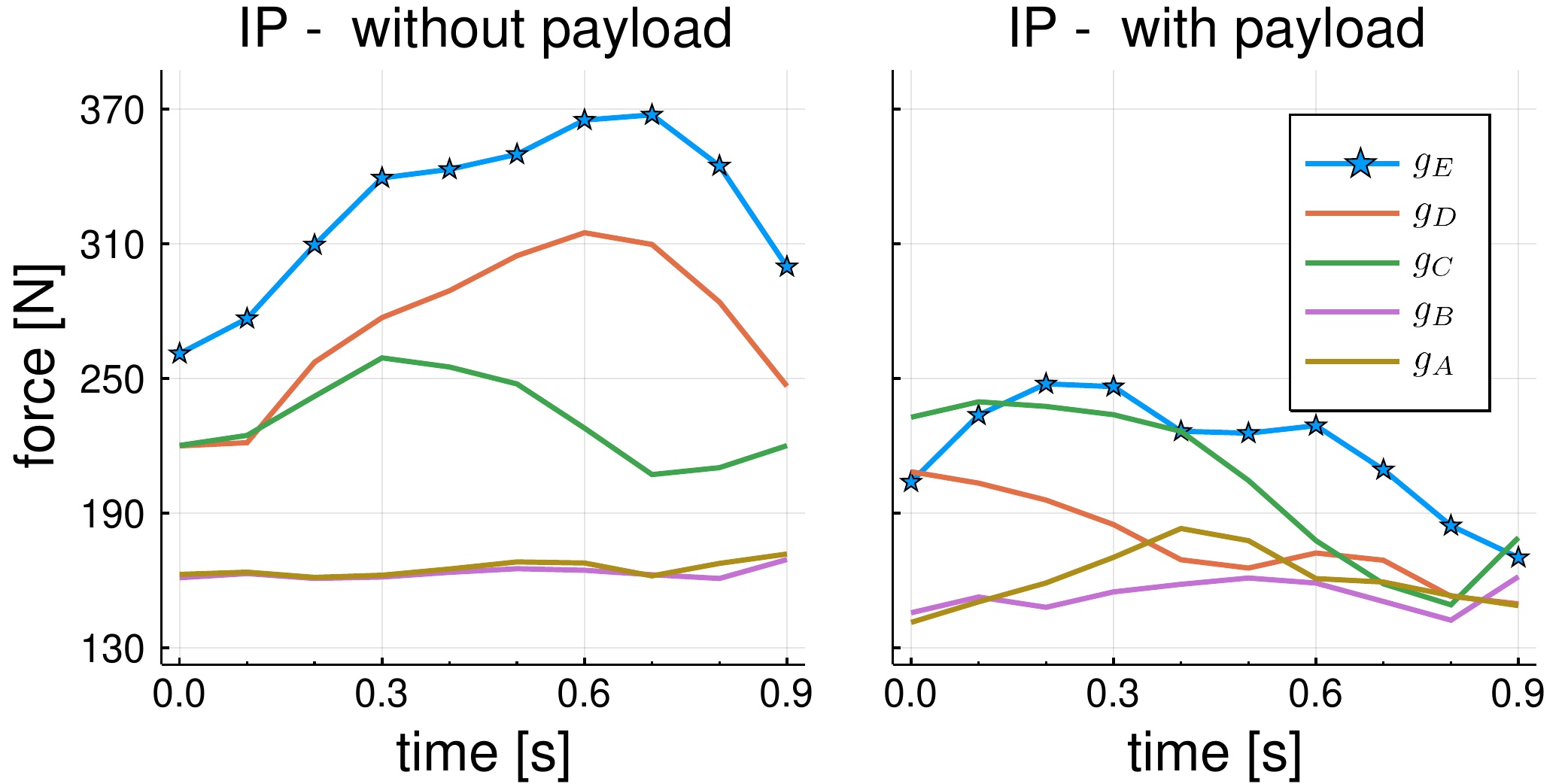}}\hfill%
    \subfigure[Sequential Quadratic Programming method.]{\includegraphics[width=0.48\linewidth]{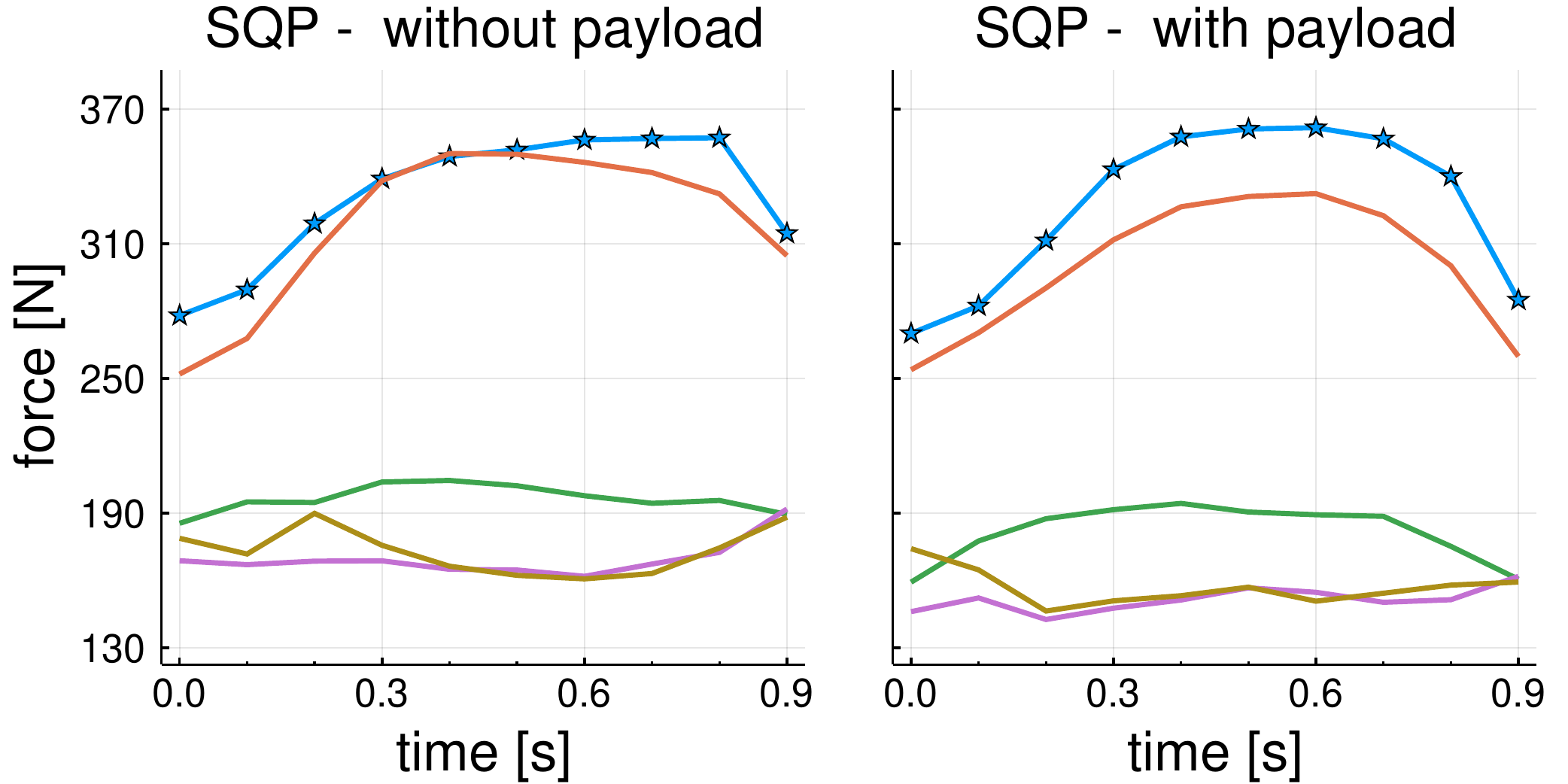}}\hfill%
    \caption{
        These plots show the maximum admissible force magnitudes over time of trajectories computed using objective functions $g_A$--$g_E$.
        We can see that the objective function $g_E$, which uses the \textit{residual force polytope}, resulted in greater admissible magnitudes than any other objective function.
    }\label{figure:suf_all}
\end{figure*}

\section{Experimental Results}
\label{sec:experimental_results}
Using a \textit{KUKA LWR} robot arm with 7-DoF, we solved the optimization problem formulated in the previous section for each of the objective functions $g_A$--$g_F$ without changing the problem constraints.
We considered \SI{1}{\second} trajectories divided into 10 equally spaced segments (11 mesh points).
\autoref{figure:trajectories} shows the motion trace of the resulting trajectories.
We used Julia~\cite{bezanson2017julia} to implement our trajectory optimization framework, and the library Knitro~\cite{byrd2006knitro} to solve the nonlinear optimization problems.

In this section, we first compare the performance of two state-of-the-art optimization methods for solving the problem we formulated.
Afterwards, we compare the obtained trajectories against each other in terms of their robustness, simulated torque expenditure, and computation time.
All evaluations were carried out in a single-threaded process on an Intel i7-6700K CPU at \SI{4.0}{\giga\hertz} and with \SI{32}{\giga\byte} \SI{2133}{\mega\hertz} memory.

\subsection{Interior-Point vs. Active-Set Methods}
We want to compare the objective functions in our formulation using different classes of optimization algorithms.
There are two broad classes of methods for solving constrained nonlinear optimization problems categorized based on how they handle constraints: %
\emph{interior-point} (IP) methods incorporate the constraints into the objective (e.g., via a barrier function or an augmented Lagrangian), while \emph{active-set} methods formulate a tractable model (e.g., by linearizing part of the constraints and penalizing them as well in the objective, as done with \gls{SQP} algorithms).
For highly nonlinear problems, SQP methods are known to suffer from excessive pivoting requiring expensive gradient evaluations of the constraints to update the active-set.
As such, they are said to scale poorly to systems with a large number of constraints. %
As a result, in robotics literature, IP methods are commonly used for direct transcription and collocation \cite{stouraitis2018dyadic,winkler2018gait}, while some rely on SQP-based solvers \cite{posa2014direct}.
However, few related work compare IP and SQP methods for solving equivalent problems, with the notable exception of \cite{winkler2018optimization}.  %
In this subsection, we compare the performance of state-of-the-art, commercial large-scale sparse IP and SQP methods on the equivalent direct transcription problem \eqref{equation:nlp} for all objective functions $g_A$--$g_F$.
This emphasizes the differences between classical IP and SQP for direct transcription applications.

We used the SQP and IP method provided by \cite{byrd2006knitro}.
For all comparisons and either method, we used automatic differentiation to obtain the Jacobian of the constraints, finite-differencing for the gradients of the objectives, and L-BFGS\footnote{L-BFGS stands for Limited-memory quasi-Newton BFGS.} for Hessian approximation (with 10 limited memory pairs).
The results are presented in \autoref{figure:iter_obj_feas}, \autoref{table:times_convergence}, and \autoref{table:ip_sqp_constraints_fevals_gevals}.
In \autoref{figure:iter_obj_feas}, we can see that the interior-point method required very few iterations to converge when compared with the active-set method.
As shown in \autoref{table:times_convergence}, the total amount of time taken to find a locally optimal solution by the interior-point method was significantly less than the active-set method.
In \autoref{table:ip_sqp_constraints_fevals_gevals}, we can see that the active-set method required significant more function and gradient evaluations than the interior-point method for the majority of the objective functions, which is expected and related to SQP's excessive pivoting (clearly observable in the feasibility error plots of SQP in \autoref{figure:iter_obj_feas}).

\subsection{Robustness to External Disturbances}
\label{subsec:robustness_to_external_disturbances}
We want to evaluate each trajectory's ability to counteract external forces while executing its planned motion.
As such, we first consider the torques required by the planned motion, and then calculate the set of all admissible forces from the remaining torques available.
We define our evaluation metric as the magnitude of the maximum admissible force considering all possible force directions.
Therefore, each trajectory is evaluated as follows:
for each point, (i) compute the residual force polytope,
then (ii) find the largest ball centered at the end-effector inscribed in that polytope, and (iii) take the radius of the ball as the robustness metric.
This is how we computed the forces shown in \autoref{figure:suf_all}.

\subsubsection{Overview of all objective functions}
\autoref{figure:suf_all} shows the evaluation results considering all objectives $g_A$--$g_E$.
In the plot, greater values correspond to greater robustness against unpredicted forces.
The trajectory computed with the \textit{residual force polytope} resulted in greater robustness than any other objective function considered.\footnote{%
    The trajectory computed with $g_C$ for the scenario with the payload and using the interior-point method resulted in an initial configuration with greater robustness than the other objective functions.
    However, we are interested in the robustness overall during the trajectory (area under the curve) and, for that, the objective function $g_E$ defined as a function of the \textit{residual force polytope} performed best.
}

\subsubsection{Force Polytope vs. Residual Force Polytope}
\begin{figure}[ht]
    \centering
    \includegraphics[width=\linewidth]{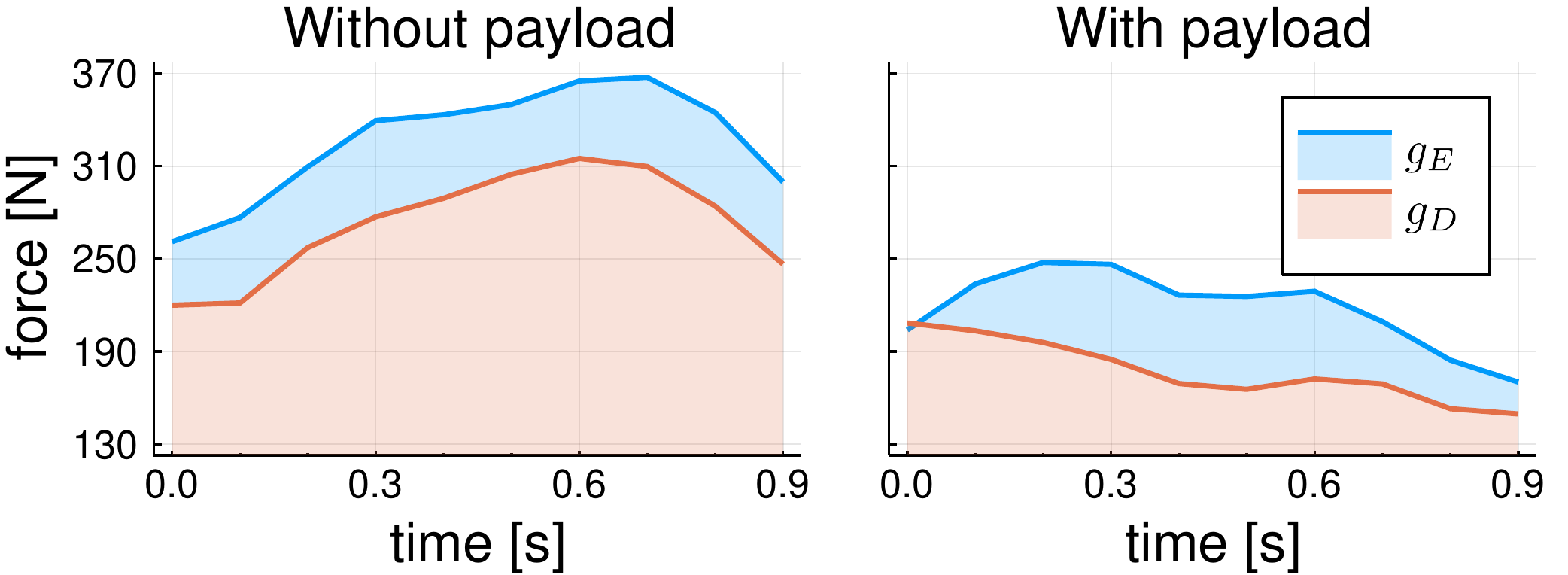}
    \caption{
        These plots show the magnitude of forces applied to the end-effector from any given direction which the robot is able to cancel out given its actuation limits.
        The solid lines represent the maximum admissible magnitude over time, and the shaded areas represent the magnitudes in between zero (no disturbance at all) and the maximum admissible magnitude.
        We can see that using the \textit{residual force polytope} (shown in blue) provided a significant improvement over the classical force polytope (shown in red).
    }\label{figure:suf_fp_rfp}
\end{figure}

\autoref{figure:suf_fp_rfp} shows the evaluation results for a scenario without a payload and for a scenario with a \SI{2}{\kilogram} cylindrical payload.
The results in the plot correspond to trajectories obtained using $g_D$ and $g_E$.
We can see that the objective using the \textit{residual force polytope} provided a significant improvement over the traditional force polytope;
more specifically, for the 1-second long trajectories we computed, an improvement of $53.2 \pm 6.52\ \si{\newton}$ without payload, and $40.55 \pm 21.37\ \si{\newton}$ with the \SI{2}{\kilogram} payload.

\subsection{Unexpected Forces vs. Expected Forces}
\begin{figure}[ht]
    \centering
    \includegraphics[width=\linewidth]{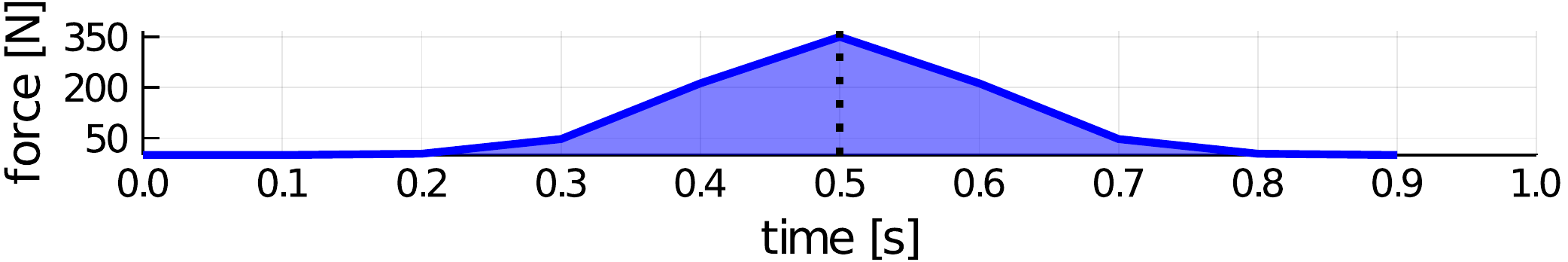}
    \caption{
        Profile of the test force applied to the end-effector.
        The impulse of this force is \SI{87.73}{\newton\second} and the peak magnitude is \SI{350}{\newton} (at $t = \SI{0.5}{\second}$).
    }\label{figure:force_profile}
\end{figure}

In this experiment, we want to compare the torque required by the trajectories optimized using objectives $g_E$ and $g_F$, which optimize a motion for resisting forces from any given direction and from a specific direction, respectively.
More specifically, we want to determine how much torque the robot would need to complete a planned motion while, at the same time, resisting an external force applied to its end-effector.
In order to do that, we apply an impulse to the robot and, for each point of the trajectory, we compute the extra torques required to oppose the external force with equation \eqref{equation:forces_from_torques}.
The magnitude of the force applied to the robot at each instant is given by
$\bm{f}(t) = \bm{f}_\mathrm{peak} \cdot \exp(-(t - 0.5)^2 / 0.02)$,
where $\bm{f}_\mathrm{peak}$ defines the magnitude at the peak of the impulse.
The profile of this test force is shown in \autoref{figure:force_profile}.

\begin{figure}[ht]
    \centering
    \includegraphics[width=\linewidth]{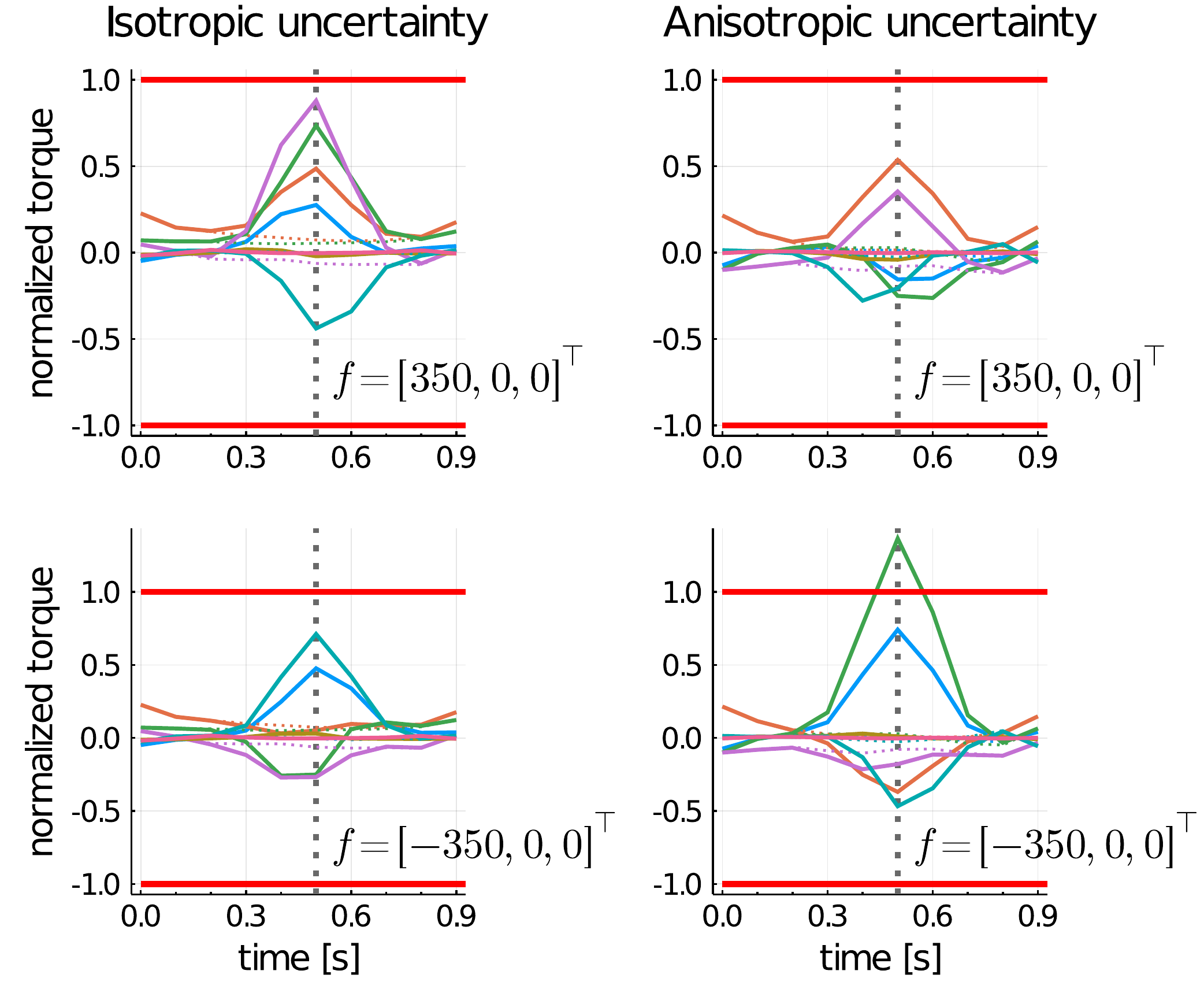}
    \caption{
        Joint torques required to complete the planned task and resist the disturbance.
        The torques have been normalized to $[-1,1]$ according to actuation limits (solid red horizontal lines).
        The nominal torques are shown as dotted lines.
        The left and right columns correspond to the trajectories computed with $g_E$ and $g_F$, respectively.
        On the left, we can see that the limits were not exceeded.
        On the top right, the trajectory resisted the impulse with less torque than $g_E$ (this impulse was applied in the same direction as the estimation during optimization).
        On the bottom right, when we applied the impulse in the opposite direction to what the specialized trajectory expected, the torques required exceeded the actuation limits of the robot.
    }\label{figure:experienced_torques}
\end{figure}

In order to compare optimal resistance to forces from any given direction ($g_E$) against optimal resistance to forces from a specific direction ($g_F$), we compute the torque required by the optimized trajectories for a test impulse that matches the direction estimation used during optimization of the specialized trajectory with $g_F$.
Afterwards, we invert the direction of the impulse and repeat the test to compute the required torques again.
The results are shown in \autoref{figure:experienced_torques}.

\subsection{Summary of Computational Runtime}
\autoref{table:objective_feval_times} shows the average time required to evaluate each of the objective functions per solver iteration.
The average was calculated from 10 samples.
It is clear that the objectives defined as functions of polytopes take significantly longer to evaluate than the other objective functions tested.

\begin{table}[ht]
    \small
    \centering
    \caption{Time required to evaluate each objective function once.}
    \label{table:objective_feval_times}
    \begin{tabular}{cS[table-format=3.3]@{\,\,\( \pm \)}S[table-format=2.3]}
        \toprule
        Objective & \multicolumn{2}{c}{Average time (\si{\milli\second})} \\ \midrule
        $g_A$     &   0.007 &  0.029 \\
        $g_B$     &   0.012 &  0.002 \\
        $g_C$     &   0.023 &  0.096 \\
        $g_D$     &  85.004 & 10.342 \\
        $g_E$     &  72.565 &  7.209 \\
        $g_F$     & 102.036 &  9.069 \\
        \bottomrule
    \end{tabular}
\end{table}

\autoref{table:times_operations} shows the average time required to compute: a force polytope, a residual force polytope, the largest ball inscribed in a polytope, the intersection of two polytopes, and the volume of a polytope.
These methods are considerably expensive and are the reason why objectives $g_D$--$g_F$  take so much time to evaluate.

\begin{table}[ht]
    \small
    \centering
    \caption{Time benchmark of computational geometry methods.}
    \label{table:times_operations}
    \begin{tabular}{lS[table-format=4]@{\ \( \pm \)\ }S[table-format=4]}
        \toprule
        Operation               & \multicolumn{2}{c}{Time (\si{\micro\second})}        \\
        \midrule
        Force polytope          & 22                                            & 124  \\
        Residual force polytope & 24                                            & 146  \\
        Largest inscribed ball  & 7354                                          & 2405 \\
        Polytope intersection   & 7081                                          & 2049 \\
        Polytope volume         & 6312                                          & 1923 \\
        \bottomrule
    \end{tabular}
\end{table}

\section{Discussion}
\label{sec:discussion}
Our initial hypothesis was that optimizing trajectories with an objective defined as a function of admissible forces in task-space---after accounting for the torques required by the motion itself---would result in motion plans more robust to external disturbances.
We defined an objective function based on the \textit{residual force polytope} to optimize a trajectory robust to forces from any given direction, and compared it against other objective functions commonly used in trajectory optimization, such as torque minimization, and manipulability maximization.
The results we obtained support our initial hypothesis:
as shown in \autoref{figure:suf_all}, for both the interior-point and active-set methods tested, the objective function $g_E$ we propose leads to optimal trajectories that are able to counteract forces from any direction with greater magnitude than any other objective function we explored.
Moreover, the objective function $g_F$, which optimizes trajectories specialized in specific directions, leads to even more robust motion plans than $g_E$ if the disturbance is applied approximately in the same direction as the one considered for the specialization.
However, specialized trajectories are less robust if the direction taken into account during optimization does not match the actual disturbance accurately (case shown in the bottom right plot of \autoref{figure:experienced_torques}).
Therefore, in terms of robustness, if a disturbance originating from a completely unexpected direction is not out of question, the objective considering any given direction ($g_E$) should be preferred over the optimization of a specific direction ($g_F$).
On the other hand, any accurate bias about disturbance directions that may arise out of known environmental  constraints (e.g., axis of fixation of articulated objects being manipulated) should be incorporated into $g_E$ to allow more dynamic range of motion.

Despite the promising results in terms of robustness, the objective functions we proposed are very demanding computationally:
even though we used a coarse problem discretization, all the objectives defined as functions of polytopes took at least 3 orders-of-magnitude longer to converge than the simpler objective functions $g_A$--$g_C$.
This significant difference is due to the double description method required to convert across polytope representations as discussed in \autoref{subsec:polytopes_and_the_double_description_method}, and due to the other mathematical operations involving polytopes (benchmarked in \autoref{table:times_operations}).
Nonetheless, the objectives $g_E$ and $g_F$ utilizing the residual force polytope representation did not incur significant convergence time differences compared to objective $g_D$ using the traditional force polytope.

\subsection{On the Scalability of Our Metric}
We did not carry out experiments using different robot arms.
While the absolute values shown in our results will vary across different manipulators, we speculate that the relative differences observed should generalize to manipulators of different sizes and with more or less joints.

Regarding the scalability of our approach to floating-base robots---such as quadrupeds or bipeds---there is a distinction to be made: whether the metric is to be used as an evaluation metric for existing trajectories, or if it is to be used as an objective function in a trajectory optimization setting.

\textbf{Robustness as an evaluation metric.}
Given an existing dynamic trajectory, computing the residual force polytope for each point in time is straightforward.
A possible application for this is to evaluate the robustness of different trajectories, and to compare them against each other.
In fact, this is exactly what we did in \autoref{subsec:robustness_to_external_disturbances} in order to evaluate the robustness of the motions obtained from the optimization of different objective functions.
For this use-case, our metric should be scalable to different platforms, but it will become more computationally demanding---and therefore slower---as the degrees of freedom of the system increase: the number of vertices of the polytope grows with the number of degrees of freedom of the system, and the complexity of converting representations (from $\mathcal{V}$-rep to $\mathcal{H}$-rep, or vice-versa) grows with the number of vertices.

\textbf{Robustness as an objective function.}
In the context of trajectory optimization, using our metric as an objective function for floating-base systems with many degrees of freedom is not straightforward and presents significant scalability issues.
The reason for this is related (but not limited) to the point mentioned above: computing the residual force polytope becomes more demanding and slower as the number of degrees of freedom of the robot increases.
For purposes of evaluating a trajectory, the polytope only needs to be computed once for each mesh point.
On the other hand, in trajectory optimization, the solver takes several iterations (in our case, hundreds of iterations) while converging to a locally optimal solution, and for each of those iterations it may need to perform more than one function or gradient evaluation, which requires computing the residual force polytope again and again.
As a consequence, optimizing trajectories for high-DoF robots in a reasonable amount of time is not possible, and could take multiple days to complete.
We would like to emphasize that this is not a limitation of the residual force polytope we propose, but a limitation of using any polytope.
Since this is a well-known issue, other authors have tried to use approximations to work around it.
Next, we list a few options for mitigating this drawback.

\subsection{Mitigating the Computational Cost}

\textbf{Less frequent polytope evaluations.}
One way to decrease computational cost is by evaluating the polytope less frequently.
For online planning and control, this would mean computing the polytope at regular time intervals, using it to adapt the motion of the robot every now and then.
This approach was used by Orsolino \textit{et al.}~\cite{orsolino2018application} for optimizing the center-of-mass position of a quadruped's static crawl gait.
In that work, the feasibility polytope was calculated once at the beginning of the optimization and used as a constant approximation thereafter.

\textbf{Approximation of polytope geometries.}
Another way to decrease computational cost is to approximate polytope geometry with morphing techniques or with surrogate models.
Bratta \textit{et al.}~\cite{bratta2020hardware} used polytope morphing for computing the polytope of each leg of a quadruped robot.
The authors computed an exact polytope representation for two key configurations, and then approximated the polytope for intermediate configurations by interpolating its shape.
Another option (not yet explored) is to use a surrogate model.
Surrogate models approximately mimic the behavior of functions that are computationally expensive to evaluate.
They can be constructed offline by exploring the states of the system, and then evaluated online quickly.

\textbf{Specialized solvers.}
In this work, we used an off-the-shelf optimization library, Knitro~\cite{byrd2006knitro}, which implements state-of-the-art algorithms for solving numerical problems.
Instead of using generic solvers, one could attempt to take advantage of problem-specific features to customize the solver's internal implementation (e.g., with heuristics, linearized relaxations, cutting planes), trading off generality for performance.
However, developing such custom solvers is very time-consuming and requires expert knowledge in numerical optimization.

\section{Conclusions}
\label{sec:conclusions}
In this paper, we proposed an exact representation for task-space forces which the robot can counteract: the \textit{residual force polytope}.
The representation takes into account the whole-body dynamics of the robot, and considers only the torques remaining after accounting for the controls of a nominal trajectory (or the controls of a trajectory being optimized).
Our proposition contrasts with approximate representations (e.g., in ellipsoidal forms) from previous related work, which do not account for the nominal control trajectory and therefore overestimate the true capabilities of a system.
We defined two functions based on the residual force polytope, for two different levels of disturbance uncertainty, and used them as objectives in trajectory optimization to plan motions more robust to external disturbances.

Despite the qualitative benefits of the trajectories obtained using our method, its computational cost does not allow deploying it as a real-time planning method.
On the other hand, our approach could be used for offline planning (where time consumption is not as critical), as well is in other areas besides trajectory optimization, such as system analysis and co-design.
Finally, addressing the long computation times required by polytope-based methods is an interesting direction for future work.
In this paper, we used explicit polytope descriptions, but with recent work from \cite{zhen2018computing} it may be possible to use approximate descriptions\footnotemark\ to considerably decrease the computational cost of planning methods using polytopes.
For that, choosing the right level of approximation becomes an important decision, and the trade-off between speed and accuracy will need to be investigated carefully.

\footnotetext{%
    Not to be confused with approximations to dynamic quantities of the controlled system, which our proposed representation is trying to avoid.
}

\bibliography{ref.bib}

\end{document}